\documentclass{article}

\usepackage[main, final]{neurips_2026}

\usepackage[utf8]{inputenc} 
\usepackage[T1]{fontenc}    
\usepackage{hyperref}       
\usepackage{url}            
\usepackage{booktabs}       
\usepackage{amsfonts}       
\usepackage{nicefrac}       
\usepackage{microtype}      
\usepackage{xcolor}         
\usepackage[utf8]{inputenc} 
\usepackage[T1]{fontenc}    
\usepackage{hyperref}       
\usepackage{url}            
\usepackage{booktabs}       
\usepackage{amsfonts}       
\usepackage{nicefrac}       
\usepackage{microtype}      
\usepackage{xcolor}         

\usepackage{latexsym}
\usepackage{amssymb}
\usepackage{amsmath}
\usepackage{amsthm}
\usepackage{booktabs}
\usepackage{enumitem}
\usepackage{graphicx}
\usepackage{color}
\usepackage{graphicx}%
\usepackage{multirow}%
\usepackage{amsmath,amssymb,amsfonts}%
\usepackage{amsthm}%
\usepackage{mathrsfs}%
\usepackage[title]{appendix}%
\usepackage{xcolor}%
\usepackage{textcomp}%
\usepackage{manyfoot}%
\usepackage{booktabs}%
\usepackage{algorithm}%
\usepackage{algorithmic}%
\usepackage{listings}%
\usepackage{bm}%
\usepackage[disable]{todonotes}%
\usepackage[utf8]{inputenc}
\usepackage{mathtools}
\usepackage{yhmath}
\usepackage{multirow}
\usepackage{wrapfig}
\usepackage{algorithm}

\usepackage{comment}

\newtheorem{theorem}{Theorem}[section]

\newtheorem{proposition}[theorem]{Proposition}%

\newenvironment{proof*}[1][\mathbb{P}oofname]{\mathbb{P}oof[#1]}
{\endproof}

\newenvironment{continuedproof*}[1]
  {
   \begin{proof}[Proof of #1 (continued)]}
  {\end{proof}}

\usepackage{mdframed}

\newmdenv[
  skipabove=\topsep,
  skipbelow=\topsep,
  linewidth=0.8pt,
  linecolor=black,
  roundcorner=4pt,
  innertopmargin=0.8\baselineskip,
  innerbottommargin=0.8\baselineskip,
  innerleftmargin=10pt,
  innerrightmargin=10pt,
]{auxproofbox}

\newif\ifuseold
\useoldfalse   

\title{
Calibrated Target Noise \\ Recovers Curvature from the Gradients
}
\author{%
  \parbox[t]{0.45\textwidth}{%
    \centering\normalfont
    \textbf{Arash Jamshidi}\\
    Department of Computer Science\\
    University of Helsinki\\
    Helsinki, Finland\\
    {\small\texttt{arash.jamshidi@helsinki.fi}}
  }%
  \And
  \parbox[t]{0.45\textwidth}{%
    \centering\normalfont
    \textbf{Katsiaryna Haitsiukevich}\\
    Department of Computer Science\\
    University of Helsinki\\
    Helsinki, Finland\\
    {\small\texttt{katsiaryna.haitsiukevich@helsinki.fi}}
  }%
  \AND
  \parbox[t]{0.45\textwidth}{%
    \centering\normalfont
    \textbf{Aristides Gionis}\\
    Division of Theoretical Computer Science\\
    KTH Royal Institute of Technology \\
    Digital Futures \\
    Stockholm, Sweden\\
    {\small\texttt{argioni@kth.se}}
  }%
  \And
  \parbox[t]{0.45\textwidth}{%
    \centering\normalfont
    \textbf{Kai Puolam{\"a}ki}\\
    Department of Computer Science\\
    Institute for Atmospheric and Earth System Research\\
    University of Helsinki\\
    Helsinki, Finland\\
    {\small\texttt{kai.puolamaki@helsinki.fi}}
  }%
}

\begin{document}

\maketitle

\begin{abstract}

Estimating curvature information from gradients alone is a fundamental problem in machine learning. The challenge becomes particularly acute in settings where only aggregate mini-batch gradients are available, yet curvature information is still required for applications such as preconditioning in optimisation, sampling, and early stopping. Perhaps surprisingly, we show that a simple target-perturbation scheme suffices to recover such curvature information from gradients alone. Our method injects zero-mean noise into the targets, with variance calibrated to the local output curvature. The covariance of the resulting noisy batch gradients then recovers practical curvature matrices, without requiring per-sample gradients, second-order derivatives, or other internal network quantities. For a broad class of non-linear models, our scheme recovers the Generalised Gauss--Newton (GGN) matrix; for feed-forward ReLU networks, it additionally recovers the diagonal blocks of the population Hessian. We demonstrate that the resulting estimator is simple, lightweight, provably optimal, and easy to integrate into existing training pipelines. Experiments demonstrate accurate curvature recovery, faster optimisation when the estimator is used for preconditioning, and the utility of the estimated GGN for early stopping.

\end{abstract}

\section{Introduction}

Modern machine-learning models are usually trained with first-order optimisation methods, yet much of their behaviour is governed by second-order structure. Curvature determines how quickly gradient-based methods can move in different directions, which directions are ill-conditioned, and how sensitive the loss is to perturbations of the parameters. As a result, curvature information is useful for preconditioning, adaptive optimisation, uncertainty estimation, sampling, and early stopping \citep{martens2010deep,mahsereci2017early}. If reliable curvature estimates were cheaply available, they could be used to make training more stable, more interpretable, and often faster.

The difficulty is that curvature is expensive to obtain in modern models. Exact Hessians are typically too large to form or store, and even practical surrogates such as the Generalised Gauss--Newton (GGN) matrix often require per-sample gradients, Jacobians, or access to intermediate model quantities \citep{martens2015optimizing,botev2017practical}. These requirements can be impractical in large-scale training pipelines, distributed systems, federated learning, privacy-constrained settings, or software frameworks that provide only aggregate mini-batch gradients. 
In such settings, the usual alternative is to ignore curvature, rely on diagonal adaptive methods, or use approximations that require more than aggregate backwards-pass information, such as per-sample gradients, Jacobians, or second-order derivatives.

This paper asks whether useful curvature information can be recovered from what is already available in such settings: aggregate mini-batch gradients. At first glance, this seems difficult, since averaging gradients over a batch destroys the per-sample structure on which curvature estimators usually rely. Moreover, the covariance of ordinary gradients does not, in general, coincide with the curvature matrix of interest. Thus, simply collecting clean gradients is not enough: the gradients must be probed in a way that makes their fluctuations carry the right geometric information.

A closely related line of work shows that noise injection can reveal hidden information in gradients. In particular, \citet{jamshidi2026information} shows that adding target noise allows the covariance of noisy gradients to recover curvature information in linear regression. However, their construction is specific to the linear-regression setting, where the Hessian coincides with the unweighted data covariance. In general, non-linear models, the relevant curvature is parameter-dependent, so target noise must be calibrated to the local geometry of the loss.

We propose the \emph{Calibrated Noisy Gradient Covariance} (\textbf{CNGC}) matrix, a simple curvature estimator based on target perturbations. The idea is to add zero-mean noise to the targets, with variance calibrated to the local curvature of the loss, and then estimate curvature from the covariance of the resulting noisy mini-batch gradients. This construction turns standard gradient evaluations into curvature probes, while avoiding explicit second-order derivatives, per-sample gradients, and Jacobian computations.

CNGC recovers the Generalised Gauss--Newton matrix in expectation for general non-linear models, and for feed-forward ReLU networks, it also captures diagonal Hessian blocks. The method is lightweight, compatible with standard mini-batch training, and can be used either as a diagnostic estimator of curvature or as a preconditioner inside an optimiser.

\textbf{Contributions.} First, we introduce CNGC as a gradient-only method for estimating curvature from calibrated target noise. Second, we show that its noisy-gradient covariance recovers the GGN matrix, and for ReLU networks also recovers diagonal Hessian blocks. Third, we analyse centred and swamping variants of the estimator and characterise their bias. Fourth, we show that scaled Rademacher noise is an optimal perturbation choice for reducing the estimator's noise-induced risk. Finally, we empirically demonstrate that CNGC accurately recovers curvature and performs favourably compared to the baselines. Additionally, we provide two versions of preconditioners based on CNGC, which improve optimisation convergence. We also show that the proposed estimator can be applied for early stopping. Our code is available at \url{https://github.com/edahelsinki/cngc}.

\section{Related work}

\paragraph{Noise injection and randomness.}
Our method uses randomness as a \emph{measurement device}: calibrated target noise is chosen so that the covariance of noisy gradients recovers a curvature target. This is conceptually related to randomised numerical linear algebra, where randomness is used to expose or preserve useful matrix structure \citep{derezinski2024recent,xu2016sub,yang2018weighted,derezinski2021newton}, but there the object being sketched is already available, whereas here curvature is made observable through first-order gradients. It is also related to SGD and its variants \citep{robbins1951stochastic,goodfellow2016deep}, whose stochasticity is known to interact with curvature and generalisation \citep{wei2019noise,zhu19anisotropic,wu2020on,xie2021diffusion,zhang2026superlinear}; the difference is that SGD noise is incidental, while ours is deliberately calibrated to the local output curvature. CNGC is closest to label-noise injection methods \citep{blanc2020implicit,haochen2021shape,damian2021label}: both perturb labels, but prior work typically uses small noise as a regulariser, whereas we scale the variance with curvature and batch size so that the gradient covariance recovers the GGN in expectation.
In \citep{jamshidi2026information}, the variance of added noise is also high, similar to this work; however, the variance of added noise is constant, and their results rely on a restrictive linearity assumption, while our work calibrates the noise to \(A''(f_w(x))\) and is applicable to general non-linear models. 
Noise can also be injected in parameter space \citep{orvieto23explicit,dauphin2024neglected,zhang2024noise}; unlike those methods, which probe the landscape through weight perturbations, we perturb targets and recover curvature from ordinary gradient evaluations. Finally, \citet{medvedev2026positive} also treat noise as beneficial, but  via input distribution perturbations rather than curvature estimation via target perturbations.

\paragraph{Recovering second-order information from gradients.}
Our work is also related to methods that extract second-order information (e.g., GGN) from first-order gradients.  Gradient-based signals have been used for optimisation monitoring and early stopping \citep{mahsereci2017early,forouzesh2021disparity,jamshidi2025}; our use of CNGC for early stopping is in this spirit, but replaces clean-gradient covariance surrogates with a curvature estimator calibrated to recover the GGN. Diagonal adaptive methods such as RMSProp, Adadelta, and Adam \citep{tieleman2012rmsprop,zeiler2012adadelta,kingma2014adam} also build inexpensive preconditioners from gradient statistics, while Newton-type, cubic-regularised, and natural-gradient methods use Hessian- or Fisher-type information to accelerate optimisation \citep{broyden1970convergence,nesterov2006cubic,amari1998natural,pascanu2014revisiting}. Our contribution is complementary: rather than proposing a new generic second-order update rule, we provide a way to \emph{estimate} the optimisation-relevant curvature from aggregate gradients alone. This also distinguishes our approach from scalable curvature approximations based on Hessian-vector products, Jacobians, Fisher approximations, or structured matrix factorisations \citep{martens2010deep,byrd2011use,martens2015optimizing,martens2018kronecker,gupta2018shampoo,zhang2023eva}. Those methods and ours share the goal of making second-order information practical, but our information model is different: variance-calibrated target perturbations make the GGN observable through aggregate mini-batch gradients. Theoretical studies such as \citet{zhao2024theoretical} characterise the Gauss--Newton matrix itself, whereas we study how to recover it from first-order observations. The closest optimisation-side comparison is \citet{liu2024sophia}: both approaches seek low-cost second-order information from stochastic gradients, but they resample labels from the model output, whereas we add zero-mean noise directly to the labels. Moreover, they give full algorithmic details only for multi-class classification. Note that the GGN also has applications beyond optimisation: combined with prior curvature, it serves as an approximate posterior precision in Laplace inference, supporting uncertainty estimation and approximate posterior sampling \citep{daxberger2021laplace}. GGN-based Laplace approximations also enable marginal-likelihood-based hyperparameter and model selection \citep{immer2021scalable}.

\section{Theory and definitions}\label{sec:tech}
\subsection{Definitions}
We assume the data comes in i.i.d.\  batches of size $n$; more specifically, batches of the form \(B=\{(x_i,y_i)\}_{i=1}^n\) where $(x_i,y_i) \sim \mathcal{D}$, for some unknown probability distribution $\mathcal{D}$. 
Throughout the paper, we consider scalar-output models $f_w(x)$ (e.g., logits in binary classification or scalar predictions in regression) parameterised by $w \in \mathbb{R}^d$ (e.g., neural network weights).
We assume loss functions of the form
\begin{equation}
\ell(f_w(x),y)=A(f_w(x)) - y\,f_w(x) \text{~~and~~} {\cal{L}}^{(B)}(w)=\sum\nolimits_{i=1}^n\ell(f_w(x_i),y_i)/n,
\end{equation}
where $A$ is a convex function, typically the log-partition function of an Exponential Family (EF) model. This formulation covers several standard losses with different choices of $A$.\footnote{For example,
$
A(z)=z^2/2
$
gives the squared $\ell_2$ loss (up to an additive constant in $y$, which has no effect on derivatives), for which
$
A''(z)=1.
$
Choosing
$
A(z)=\log(1+e^z)
$
gives the logistic, or binary cross-entropy, loss for \(y\in\{0,1\}\), for which
$
A''(z)=\sigma(z)\bigl(1-\sigma(z)\bigr),
$
where
$
\sigma(z)=1/(1+e^{-z})
$
is the logistic sigmoid function.}
The \textit{Generalised Linear Models} (GLMs) are the special case where $f_w(x)=w^\top x$ is a linear function \citep{mccullagh2019Generalized}. An overview of GLMs and EF losses is presented in Section~\ref{prelem:glm}. 
For brevity, we write \(J_w(x):=\nabla_w f_w(x)\) for the gradient of the model output with respect to the parameters.

In non-linear models, the positive semidefinite curvature matrix most commonly used for second-order optimisation in modern neural networks is the \textit{Generalised Gauss--Newton} (GGN) matrix \citep{botev2017practical}, which is the main focus of this paper, defined for a batch $B$ and the population,
\begin{equation}\label{eq:GGN}\boxed{
G^{(B)}(w)
=
\sum\nolimits_{i=1}^{n}
A''(f_w(x_i))\,J_w(x_i)J_w(x_i)^\top/n
\text{~~and~~}
G(w)
=
\mathbb{E}_{B\sim \mathcal{D}^n}[G^{(B)}(w)].
}
\end{equation}

For comparison, the empirical and population Hessians are
\begin{equation}
H^{(B)}(w)
=
\sum\nolimits_{i=1}^{n}\nabla_w^2\,\ell(f_w(x_i),y_i)/n
\text{~~and~~}
H(w)
=
\nabla_w^2\,\mathbb{E}_{(x,y)\sim\mathcal{D}}\bigl[\ell(f_w(x),y)\bigr].
\end{equation}

A direct calculation gives the decomposition (the analogous population identity is obtained by taking expectation over \((x,y)\sim\mathcal{D}\)),
\begin{equation}
G^{(B)}(w)
=
H^{(B)}(w)
-
\sum\nolimits_{i=1}^{n}
\bigl(A'(f_w(x_i)) - y_i\bigr)\,\nabla_w^2 f_w(x_i)/n.
\end{equation}

Thus, the difference between the Hessian and the GGN is exactly the term involving the second derivative of the model output. In particular, for linear models (GLMs), where \(\nabla_w^2 f_w(x)=0\), the GGN reduces to the Hessian.
The GGN is the practical optimisation target, while the Hessian enters mainly as a comparison object and as the exact GLM special case.

Our only assumption about the distribution $\mathcal{D}$ is that all random variables in our work have finite moments specified below:
\begin{equation}\label{eq:assumption}
    \mathbb{E}\|J_w(x)\|_2^4 < \infty,
    \quad
    \mathbb{E}\Big[
        \big(A'(f_w(x)) - y\big)^2
        \|J_w(x)\|_2^2
    \Big]
    < \infty,
    \quad
    \mathbb{E}\Big[
        A''(f_w(x))
        \|J_w(x)\|_2^2
    \Big]
    < \infty.
\end{equation}

We denote the one-sample loss gradients by
\begin{equation}
g_i(w):=\nabla_w\,\ell(f_w(x_i),y_i)
=
\bigl(A'(f_w(x_i)) - y_i\bigr)\,J_w(x_i),
\end{equation}
and assume access only to aggregate batch gradients,
\begin{equation}
g^{(B)}(w)=\sum\nolimits_{i=1}^n g_i(w)/n.
\end{equation}
Our method does not require per-sample gradients, Jacobians, or second-order derivatives. It does, however, assume access to per-example model outputs and targets within each minibatch in order to compute the calibrated perturbations.
A gradient optimisation run typically yields a sequence of such gradients
\(g^{(B_1)}(w_1),\dots,g^{(B_t)}(w_t)\)
evaluated at consecutive optimisation steps \(w_1,\dots,w_t\).

\subsection{Variance-calibrated target noise}

Our approach is to perturb the targets independently by calibrated noise according to
\begin{equation}\label{eq:noiserule}
\boxed{
\tilde y_i = y_i + \sqrt{\beta}\,\xi_i,
\qquad
\mathbb{E}[\xi_i \mid x_i] = 0,
\qquad
\operatorname{Var}[\xi_i\mid x_i]=n\, A''(f_w(x_i)),
}
\end{equation}
where \(\beta>0\) controls the noise level. Let \(\tilde g^{(B)}(w)\) denote the batch gradient computed using the perturbed targets. Then
\begin{equation}
\tilde g^{(B)}(w)-g^{(B)}(w)
=
-\sqrt{\beta} \sum\nolimits_{i=1}^n \xi_i J_w(x_i)/n,
\end{equation}
and in particular, the perturbation is unbiased:
$
\mathbb{E}_\xi[\tilde g^{(B)}(w)\mid B]=g^{(B)}(w)$.

This suggests two covariance estimators to estimate the GGN. The first is the \textit{centred estimator}, which uses the clean gradient of the same batch as the centre:
\begin{equation}\label{eq:cent}
\boxed{
\widehat S^{(B)}_{\mathrm{cent}}(w)
:=
\bigl(\tilde g^{(B)}(w)-g^{(B)}(w)\bigr)
\bigl(\tilde g^{(B)}(w)-g^{(B)}(w)\bigr)^\top/\beta.
}
\end{equation}

The second is the \textit{swamping estimator}, which centres noisy gradients around their mean:
\begin{equation}\label{eq:swamped}
\boxed{
\widehat S^{(B)}_{\mathrm{swamp}}(w)
:=
\bigl(\tilde g^{(B)}(w)-\mu(w)\bigr)
\bigl(\tilde g^{(B)}(w)-\mu(w)\bigr)^\top/\beta,
\qquad
\mu(w):=\mathbb{E}_{\xi,x,y}[\tilde g^{(B)}(w)].
}
\end{equation}
The identity in Eq.~\eqref{eq:swamped} is stated for the idealised centre \(\mu(w)=\mathbb{E}_{\xi,x,y}[\tilde g^{(B)}(w)]\); in practice we replace \(\mu(w)\) by an empirical approximation, introducing additional finite-sample centring error.

Each sample \(\widehat S^{(B)}_{\mathrm{cent}}(w)\) and \(\widehat S^{(B)}_{\mathrm{swamp}}(w)\) is rank one, so recovering a higher-rank approximation to the GGN requires aggregation across multiple probes. For \(K\) independent batches \(B_1,\dots,B_K\) evaluated at the same parameter value \(w\), define
\begin{equation}\label{eq:K}
\overline S_{\mathrm{cent},K}(w)
:=
\sum\nolimits_{k=1}^K
\widehat S^{(B_k)}_{\mathrm{cent}}(w)/K
\text{~~~and~~~}
\overline S_{\mathrm{swamp},K}(w)
:=
\sum\nolimits_{k=1}^K
\widehat S^{(B_k)}_{\mathrm{swamp}}(w)/K.
\end{equation}
These aggregated estimators provide practical higher-rank approximations to the batch- or population-level GGN, respectively. \footnote{In optimisation applications, we typically do not evaluate several independent mini-batches at exactly the same parameter value $w$, since the parameters are updated after each mini-batch. In practice, we average the estimator over an epoch, which can be interpreted as a running-average approximation to the GGN along the optimisation trajectory.} 

\paragraph{Why do clean gradients fail?} The outer product of the clean one-sample gradient is
$
g_i(w)g_i(w)^\top
=
\bigl(A'(f_w(x_i)) - y_i\bigr)^2\,J_w(x_i)J_w(x_i)^\top$,
which differs from the GGN of Eq. \eqref{eq:GGN}. Thus, the covariance of clean gradients is weighted by squared residuals, while the GGN is weighted by \(A''(f_w(x_i))\). In general, these are different objects, so ordinary gradient covariance does not recover the GGN. For a batch gradient, the same mismatch remains, and the corresponding covariance is simply scaled by \(1/n\).

\paragraph{Formal statement (Eqs.~\eqref{eq:cent} and \eqref{eq:swamped} reproduce the GGN).}
The next theorem makes this precise under the moment assumptions in Eq.~\eqref{eq:assumption}. It shows that the centred estimator recovers the batch GGN exactly in conditional expectation, whereas the swamping estimator recovers the population GGN up to an explicit residual term of order \(1/(n\beta)\) that is independent of the injected-noise distribution.
\begin{theorem}\label{thm:gradcovGGN}
Assume an i.i.d.\ batch \(B=\{(x_i,y_i)\}_{i=1}^n\) satisfying Eq.~\eqref{eq:assumption}. Then the estimators defined above in Eqs.~\eqref{eq:cent} and \eqref{eq:swamped} satisfy
\begin{equation}
\mathbb{E}_\xi\!\bigl[\widehat S^{(B)}_{\mathrm{cent}}(w)\mid B\bigr]
=
G^{(B)}(w),
\qquad
\mathbb{E}_{\xi,x,y}\!\bigl[\widehat S^{(B)}_{\mathrm{swamp}}(w)\bigr]
=
G(w)+\Delta(w)/(n\beta),
\end{equation}
where \(\Delta(w)\) depends only on the data distribution \(\mathcal{D}\) and the parameter \(w\), and is independent of the injected-noise distribution. Moreover,
$
\|\Delta(w)/(n\beta)\|_{\mathrm{op}}
=
\mathcal{O}(1/(n\beta))$,
where \(\|\cdot\|_{\mathrm{op}}\) denotes the operator norm.\footnote{For a matrix \(A\in\mathbb{R}^{m\times n}\), the operator norm is defined by \(\|A\|_{\mathrm{op}}=\sup_{\|x\|_2=1}\|Ax\|_2\), which is equal to the largest singular value of \(A\).}
\end{theorem}
The proof is in Appendix \ref{sec:GGNproof}.
The theorem shows that the centred estimator is unbiased for the batch GGN and that the swamping estimator becomes increasingly accurate as either the batch size \(n\) or the noise level \(\beta\) increases. In particular, the role of \(\beta\) differs in the two constructions: for the centred estimator, it is only a scaling factor that disappears after normalisation by \(1/\beta\), whereas for the swamping estimator, it directly controls the size of the residual bias term.

\textbf{Vector outputs.}
Throughout the main development, we focus on scalar-output models mainly for simplicity of notation. However, the same variance-calibration principle extends to vector-valued outputs, where the scalar curvature term is replaced by the corresponding output-space curvature matrix, and the injected noise is required to have an appropriate covariance structure rather than a scalar variance. An important example is multi-class classification with the standard multi-class cross-entropy loss, where \(A''(\cdot)\) is replaced by the Hessian of the log-sum-exp link. Since this is a matrix-valued version of the same argument, we defer the formal statement and proof to Subsection~\ref{app:multiclass_extension}.

\subsection{Special case:  networks with ReLU non-linearities}

Theorem~\ref{thm:gradcovGGN} already applies to general scalar-output neural networks: once \(J_w(x)=\nabla_w f_w(x)\) is viewed as the feature vector, the variance-calibrated target-noise construction in Eq.~\eqref{eq:noiserule} and the estimators in Eqs.~\eqref{eq:cent}--\eqref{eq:swamped} recover the GGN of Eq.~\eqref{eq:GGN} in expectation using only aggregate batch gradients. In general, the recovered object is the GGN rather than the full Hessian.

\textbf{Having an additional backward pass.}
In many practical settings, it is preferable to have an additional backward pass with clean targets. First, the noisy gradient is often not suitable for a reliable optimisation update, since the injected target noise can be large. Second, access to the clean gradient allows us to subtract it from the noisy gradient, isolating the term of interest. This is exactly the idea behind the centred estimator in Eq.~\eqref{eq:cent}, which yields an unbiased estimator of the batch GGN.

\textbf{Feed-forward ReLU networks.}
With non-zero pre-activations, a feed-forward ReLU network is locally linear in each layer's parameters when the other layers are fixed. Each block therefore defines a local GLM, so its Hessian and GGN blocks coincide and the calibrated-noise argument applies blockwise. This result is pointwise: activation patterns may change between training steps. Target perturbations preserve the activation pattern because clean and noisy gradients use the same parameters and inputs.

\begin{proposition}\label{prop:relu_block_hessian}
Consider a feed-forward ReLU network with scalar output, and let \(u_j=\mathrm{vec}(W_j)\) denote the parameter block of layer \(j\). Assume that, at the fixed parameter \(w\), all pre-activations are non-zero. Define
\begin{equation}
H_{jj}^{(B)}(w)
:=
\sum\nolimits_{i=1}^n \nabla_{u_j}^2\,\ell(f_w(x_i),y_i)/n ,
\qquad
H_{jj}(w)
:=
\nabla_{u_j}^2\,\mathbb E_{(x,y)\sim\mathcal D}\bigl[\ell(f_w(x),y)\bigr].
\end{equation}
Let \(g_{u_j}^{(B)}(w)\) and \(\tilde g_{u_j}^{(B)}(w)\) denote the restrictions of \(g^{(B)}(w)\) and \(\tilde g^{(B)}(w)\) to the block \(u_j\), and define \(\widehat S^{(B)}_{\mathrm{cent},u_j}(w)\) and \(\widehat S^{(B)}_{\mathrm{swamp},u_j}(w)\) by Eqs.~\eqref{eq:cent} and \eqref{eq:swamped} with \(g^{(B)}(w)\), \(\tilde g^{(B)}(w)\) replaced by \(g_{u_j}^{(B)}(w)\), \(\tilde g_{u_j}^{(B)}(w)\). Then
\begin{equation}
\mathbb E_\xi\!\bigl[\widehat S^{(B)}_{\mathrm{cent},u_j}(w)\mid B\bigr]
=
H_{jj}^{(B)}(w),
\qquad
\mathbb E_{\xi,x,y}\!\bigl[\widehat S^{(B)}_{\mathrm{swamp},u_j}(w)\bigr]
=
H_{jj}(w)+\Delta_j(w)/(n\beta),
\end{equation}
where \(\Delta_j(w)\) depends only on \(\mathcal D\) and \(w\), and is independent of the injected-noise distribution.
\end{proposition}

\textbf{Remark (non-i.i.d. batches).} The centred identities in Theorem~\ref{thm:gradcovGGN} and Proposition~\ref{prop:relu_block_hessian} hold for any fixed batch and do not require i.i.d. data. The expectation is taken only over the conditionally independent, variance-calibrated target perturbations. Thus, the centred estimator remains unbiased for the batch GGN and, under the ReLU assumptions, for the batch Hessian's diagonal blocks. In contrast, the population-level bias formulas for the swamping estimator rely on i.i.d. sampling.

\textbf{Remark (block-diagonal structure).}
It is often desirable to recover only a block-diagonal Hessian structure. First, the Hessian is often dominated by its diagonal blocks \citep{martens2015optimizing,botev2017practical}. Second, restricting to diagonal blocks makes storage and inversion tractable, especially for preconditioning. Many practical second-order methods, therefore, rely on block-diagonal approximations.

The same local-linearity argument extends beyond fully connected feed-forward networks. In particular, many modern architectures, such as convolutional neural networks and graph neural networks, can be viewed as stacks of linear layers with architecture-specific weight sharing, so the same blockwise reasoning applies whenever the architecture uses piecewise linear activations \citep{eschenhagen2023kronecker}.

\subsection{Scaled Rademacher distribution is optimal}
\label{subsec:rademacher-optimal}

The target matrices in Theorem~\ref{thm:gradcovGGN} depend only on the conditional second moments of the perturbations \(\xi_i\), not on their full distribution. This raises the natural question of which admissible perturbation law should be used in practice. We show that a simple scaled Rademacher perturbation is optimal in the sense that, among all zero-mean perturbations with the required conditional variance, it minimises the noise-induced variance of the GGN estimators.
We therefore define the independent perturbations for $i\in\{1,\ldots,n\}$ by
\begin{equation}\label{eq:rademacher}
\boxed{
\xi_i = \epsilon_i \sqrt{n\,A''(f_w(x_i))},
\qquad
\epsilon_i\in\{-1,+1\},
\qquad
P(\epsilon_i=-1)=P(\epsilon_i=+1)=1/2.}
\end{equation}
By construction, Eq.~\eqref{eq:rademacher} satisfies
$
\mathbb E[\xi_i\mid x_i]=0$ and 
$\mathbb E[\xi_i^2\mid x_i]=n\,A''(f_w(x_i))$.

\begin{proposition}[Optimality of scaled Rademacher noise]
\label{prop:optimal_noise}
Fix a batch \(B\) and a parameter \(w\), and consider conditionally independent perturbations \(\xi_i\) satisfying
\begin{equation}
\mathbb E[\xi_i\mid x_i]=0,
\qquad
\mathbb E[\xi_i^2\mid x_i]=n\,A''(f_w(x_i)).
\end{equation}
Then the scaled Rademacher law in Eq.~\eqref{eq:rademacher} minimises the conditional noise-induced Frobenius risk of the centred estimator in Eq.~\eqref{eq:cent} among all such perturbation laws. If, in addition, the admissible perturbation laws satisfy
\begin{equation}
\mathbb E[\xi_i^3\mid x_i]=0,
\end{equation}
then the same scaled Rademacher law also minimises the conditional noise-induced Frobenius risk of the swamping estimator in Eq.~\eqref{eq:swamped}. The same conclusion holds for the conditional mean-squared directional error \(v^\top (\widehat S - G) v\) for every unit vector \(v\in\mathbb R^d\).
Moreover, these minimisers are the same for arbitrary \(\beta>0\).
\end{proposition}

The proposition explains why symmetric two-point perturbations are the natural default in practice: they achieve the required variance, are easy to sample, and minimise the perturbation-induced variability of the estimator. 
More generally, conditioned on any fixed batch \(B\), any admissible noise distribution with finite first four moments leads to \textit{finite perturbation-induced variability}. However, among noise distributions with the same conditional variance, the scaled Rademacher distribution has the smallest fourth moment and therefore minimises this variability. In contrast, alternatives such as Gaussian noise have larger fourth moments and therefore induce larger estimator error.

\textbf{Remark.}
For regression and binary classification, GNB probes of the form used in Sophia \citep{liu2024sophia} correspond to centred estimators with scaled Gaussian and centred Bernoulli noise, respectively. Proposition~\ref{prop:optimal_noise} therefore guarantees that our centered estimator (using scaled Rademacher perturbation) yield better  mean-squared error for recovering GGN in both regression and binary classification, compared to GNB.

\subsection{Algorithm description}
\label{subsec:algorithm-description}

Algorithm~\ref{alg:centered-swamping} gives a compact implementation of the centred and
swamping estimators. Both use the same variance-calibrated target perturbation, with
$v_i=nA''(\eta_i)$, and differ only in the centring step. The centred estimator subtracts the clean
gradient from the noisy gradient, while the swamping estimator centres noisy gradients around
$\mu\approx\mathbb{E}[\tilde g]$.
One example of using these two estimators in optimisation is discussed in Section~\ref{subsec:blockwise-preconditioned-optimisation} and Algorithm~\ref{alg:blockwise-preconditioned-centered}. Algorithm~\ref{alg:centered-swamping} uses a running empirical centre \(\mu_t\) to approximate the idealised centre \(\mu(w)\) from Eq.~\eqref{eq:swamped}, so the practical swamping estimator is an approximation to the theoretical one in Theorem~\ref{thm:gradcovGGN} rather than exactly the same object.

\begin{algorithm}[t]
\caption{Centred and swamping noisy-gradient curvature (CNGC) estimators.}
\label{alg:centered-swamping}
\centering
\small
\begin{tabular}{@{}p{0.47\linewidth}@{\hspace{0.04\linewidth}}p{0.47\linewidth}@{}}
\hline
\rule{0pt}{2.6ex}\textbf{Centered estimator} &
\rule{0pt}{2.6ex}\textbf{Swamping estimator} \\[0.8ex]
\hline
\\[-1.8ex]

1: \textbf{Input:} parameter $w$, noise multiplier $\beta$. &
1: \textbf{Input:} parameter $w$, noise multiplier $\beta$, {\color{blue}center $\mu_t$}. \\

2: Draw a mini-batch $\{(x_i,y_i)\}_{i=1}^{n}$. &
2: Draw a mini-batch $\{(x_i,y_i)\}_{i=1}^{n}$. \\

3: Compute $\eta_i=f_w(x_i)$, $\forall i\in[n]$. &
3: Compute $\eta_i=f_w(x_i)$, $\forall i\in[n]$. \\

4: Set $v_i = n A''(\eta_i)$, $\forall i\in[n]$. &
4: Set $v_i = n A''(\eta_i)$, $\forall i\in[n]$. \\

5: Sample $\xi_i\sim\mathrm{Unif}\{+\sqrt{v_i},-\sqrt{v_i}\}$, $\forall i\in[n]$. &
5: Sample $\xi_i\sim\mathrm{Unif}\{+\sqrt{v_i},-\sqrt{v_i}\}$, $\forall i\in[n]$. \\

6: Set $\tilde y_i = y_i+\sqrt{\beta}\xi_i$, $\forall i\in[n]$. &
6: Set $\tilde y_i = y_i+{\sqrt{\beta}}\xi_i$, $\forall i\in[n]$. \\

7: Compute $\tilde g=\nabla_w \sum_i \ell(f_w(x_i),\tilde y_i) / n$. &
7: Compute $\tilde g=\nabla_w \sum_i \ell(f_w(x_i),\tilde y_i) / n$. \\

8: {\color{blue}Compute $g=\nabla_w \sum_i \ell(f_w(x_i),y_i) / n$.} &
8: {\color{blue}Update $\mu_{t+1} = (t \mu_t + \tilde g)/(t+1)$.} \\

9: \textbf{return} $(\tilde g-{\color{blue}g)}(\tilde g-{\color{blue}g})^\top/\beta$. &
9: \textbf{return} $(\tilde g-{\color{blue}\mu_{t+1}})(\tilde g-{\color{blue}\mu_{t+1}})^\top {/ \beta}$. \\
\hline
\end{tabular}
\end{algorithm}

\section{Experiments}

To evaluate the proposed curvature estimator, we state the following research questions: \textbf{(i)} does the estimator recover the curvature information, and what are the effects of centring, swamping, and noise shape (Section~\ref{sec:curv_est_theory}), \textbf{(ii)} how does the proposed estimator compare to the existing methods (Section~\ref{sec:curv_est_baselines}), \textbf{(iii)} does the estimator improve the optimisation when used as a preconditioner across various architectures (Section~\ref{sec:preconditioner}), \textbf{(iv)} can the estimator provide a criterion for early stopping (Section~\ref{sec:early_stopping}). We use the relative operator norm defined as follows to evaluate the approximation:
\begin{equation}
    r = ||S_g - G||_\mathrm{op}/||G||_\mathrm{op}\,,
\label{eq:norm_r}
\end{equation}
where $S_g$ is an approximation and $G$ is either GGN or Hessian. We also evaluated the difference in (positive) eigenvalues between the estimation and the ground truth by the same metric, since the eigenspectrum describes the loss landscape curvature.

\subsection{Curvature estimators: effects of centring, swamping, and noise design} \label{sec:curv_est_theory}
Our theoretical results suggest three distinct empirical questions related to the design of the estimator:
\emph{(a) Curvature recovery and effects of centring:} does the centred same-batch estimator recover \(G^{(B)}(w)\)? Does the swamping cross-batch estimator recover \(G^{(B)}(w)\)?
\emph{(b) Swamping:} does the swamping estimator exhibit the predicted \(O((n\beta)^{-1})\) bias?
\emph{(c) Noise design:} does symmetric two-point noise dominate Gaussian noise at fixed variance, as predicted by Proposition~\ref{prop:optimal_noise}?

\begin{figure}
    \centering
    \includegraphics[width=0.25\textwidth, trim={0 7 0 6}, clip]{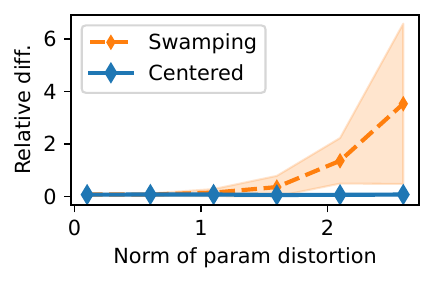}
    \includegraphics[width=0.245\textwidth, trim={16 7 0 6}, clip]{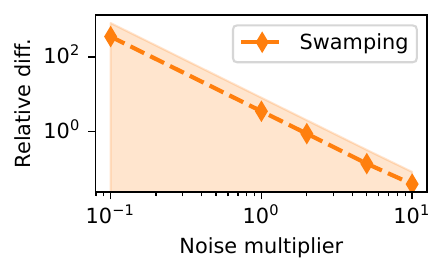}
    \includegraphics[width=0.245\textwidth, trim={16 7 0 6}, clip]{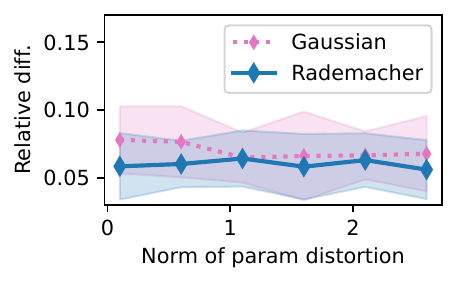}
    \caption{Relative operator norm between GGN and its estimates showing the effects of centring (\textit{left}), swamping (\textit{middle}) and the added noise distribution (\textit{right}) on curvature recovery. Both centring and swamping with a large noise multiplier make the estimates robust. Rademacher noise distribution yields a more accurate estimator with lower variance.}
    \label{fig:curv_est}
 \end{figure}

Figure~\ref{fig:curv_est} summarises the results. \textit{(a)} Both estimators recover the GGN, but centring significantly improves the robustness of the estimate far from the true parameter values (\textit{left}). \textit{(b)} Setting $\beta$, as a multiplier, we showed that the swamping estimator exhibits the predicted bias, however, by increasing $\beta$ sufficiently, the estimator yields an accurate curvature estimate (\textit{middle}). \textit{(c)} Both symmetric two-point (Rademacher) and Gaussian noise distributions lead to accurate estimates of the curvature; however, Rademacher noise exhibits lower variance, leading to a better estimator (\textit{right}). See run details in Section~\ref{sec:curv_theory}. Additionally, Section~\ref{sec:illustrative} provides an illustrative example for the Hessian recovery of a small CNN applied to regression and classification tasks.

\textbf{Recovery of Hessian eigenvalues} We further tested the calibration effects on the recovery of (positive) eigenvalues of the true Hessian (see Section~\ref{sec:h_recovery_exp} and Figure~\ref{fig:curv_est_hessian} in Appendix). Both estimators with centring and swamping with a large enough noise multiplier capture the eigenspectrum of the Hessian.

\subsection{Curvature estimators: comparison with the baselines}\label{sec:curv_est_baselines}
\begin{figure}
    \centering 
    \includegraphics[width=0.245\linewidth,trim={3 0 2 0},clip]{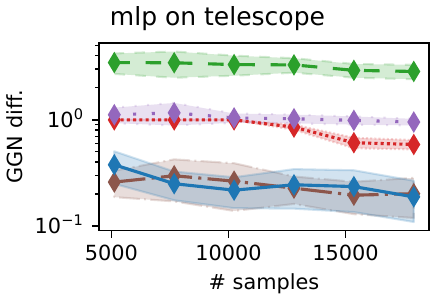}
    \includegraphics[width=0.235\linewidth,trim={13 0 2 0},clip]{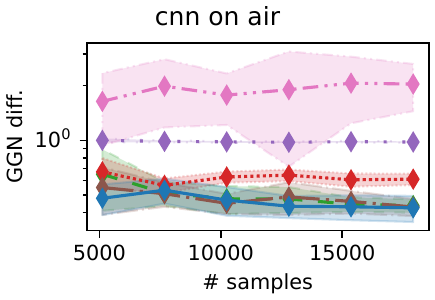}
    \includegraphics[width=0.34\linewidth,trim={3 0 2 0},clip]{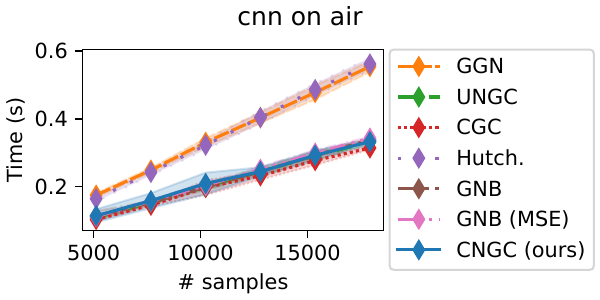}
    \caption{Comparison of GGN estimators with an MLP for a classification task (\textit{left}) and a CNN for a regression task (\textit{middle}) by the relative operator norm of the difference of the full GGN matrix and its estimators. The wall-clock time for calculation of the estimators for the CNN model (\textit{right}). The proposed CNGC estimator performs consistently across classification and regression tasks and is computationally cheaper than the full GGN calculation.}
    \label{fig:est}
\end{figure}

We calculated the GGN matrix by the definition in Eq.~\eqref{eq:GGN} as the ground truth for the evaluation of estimators. We compared the proposed estimator CNGC with the following baselines: \textbf{a)} clean gradient covariance (CGC, the covariance matrix of mean batch gradients without added noise); \textbf{b)} uncalibrated noisy gradient covariance (UNGC) \citep{jamshidi2026information} (adds target noise with constant variance); \textbf{c)} Hutchinson estimator \citep{bekas2007estimator} (evaluates only the Hessian diagonal); \textbf{d)} Gauss-Newton-Bartlett curvature estimator (samples the model outputs instead of added noise) \citep{liu2024sophia}.
We compared the estimators reporting the relative operator norm of the error in the full GGN matrix by Eq.~\eqref{eq:norm_r} and the wall-clock time of their calculation in Figure~\ref{fig:est}. The comparison between the matrix eigenspectrums for the estimators and the ground truth GGN is in Figure~\ref{fig:est_appendix}.

Neither the naive CGC nor the Hutchinson estimator fully captures the GGN. Since the Hutchinson estimator targets only diagonal structure, the full-matrix comparison is not favourable to it, however, it recovers the Hessian eigenspectrum (Figure~\ref{fig:est_appendix},  \textit{bottom left}). UNGC fails for the classification task while recovering the GGN for the regression. Similarly, GNB performs classification but requires calibration for regression (see `GNB' and `GNB (MSE)'). Additionally, GNB utilises Gaussian noise, which yields an estimator with higher variance as shown in the previous section. The proposed CNGC works for both tasks and matches the computational efficiency of the  baselines.

\subsection{CNGC estimator as the preconditioner in optimisation}\label{sec:preconditioner}
In this section, we verify if the proposed curvature estimator can be used as a preconditioner (research question \textbf{(iii)}). \textbf{Baselines}: SGD \citep{robbins1951stochastic,sutskever2013importance}, Adam \citep{kingma2014adam}, Muon \citep{jordan2024muon}
, Sophia-G \citep{liu2024sophia}. For the proposed preconditioner, we used two versions of the optimisation algorithm: the block-diagonal preconditioner (`GD, PC') with adaptive learning rate (see Alg.~\ref{alg:blockwise-preconditioned-centered} for details) and the diagonal preconditioner with the same momentum and clipping techniques as Sophia optimiser (`SophiaC'). 
We tested linear, MLP, CNN, and transformer encoder models on classification and regression tasks (see Section~\ref{sec:precond}).
For all runs, the input features were normalised to have zero mean, unit variance, and a batch size of 128. The preconditioner used the centred version of the estimator by subtracting the clean gradient. In the block-diagonal preconditioner version, the blocks were inverted using the pseudo-inverse.
The proposed method, including the preconditioner and  Alg.~\ref{alg:blockwise-preconditioned-centered}, yields results that are competitive with those of other popular optimisers.
Figure~\ref{fig:pc_cnn_transf} (and Figure~\ref{fig:pc_linear_mlp_append} in the Appendix) summarises the results. The preconditioner effectively recovers the curvature information as shown by the `GD, PC' version. Additionally, incorporating the estimator with effective clipping (`SophiaC') slightly improves the results of the original GNB estimator for regression tasks, thanks to a better noise design. 

\begin{figure}
    \centering
    \includegraphics[width=0.205\linewidth,trim={10 0 10 0},clip]{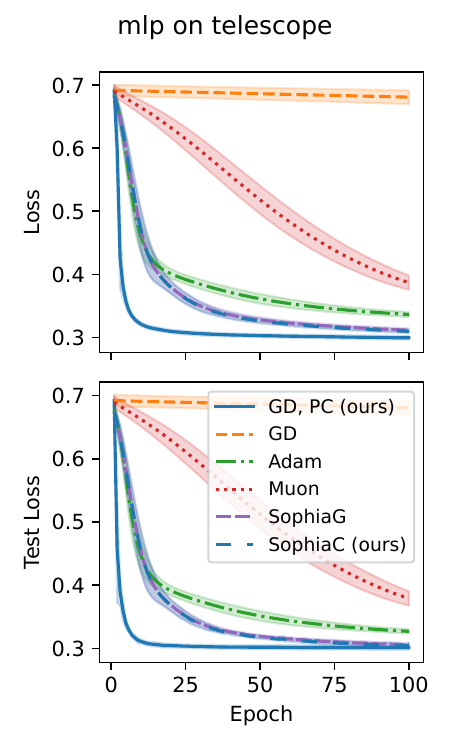}
    \includegraphics[width=0.19\linewidth,trim={25 0 10 0},clip]{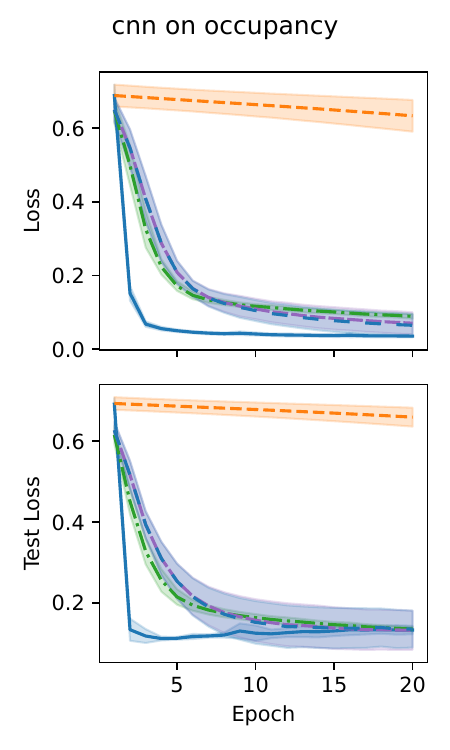}
    \includegraphics[width=0.19\linewidth,trim={25 0 10 0},clip]{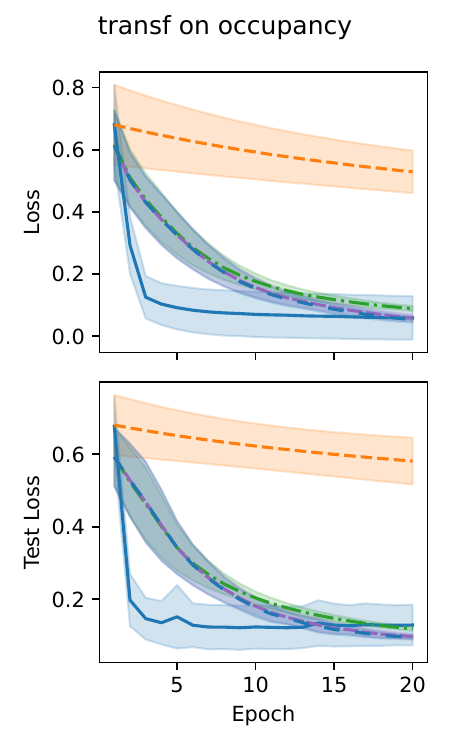}
    \includegraphics[width=0.19\linewidth,trim={25 0 10 0},clip]{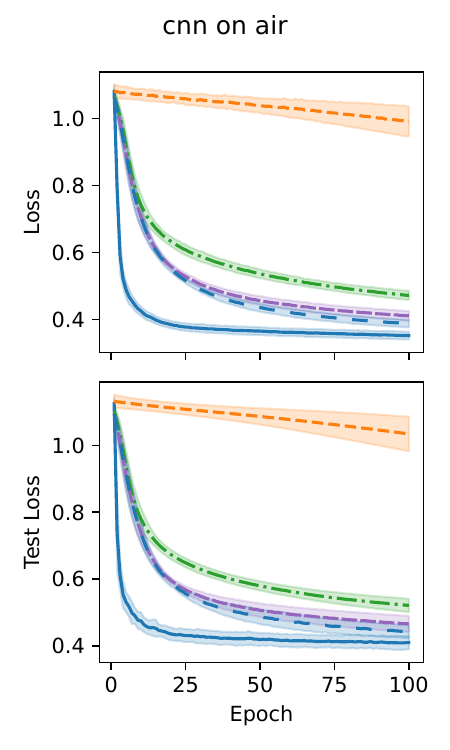}
    \includegraphics[width=0.19\linewidth,trim={25.5 0 10 0},clip]{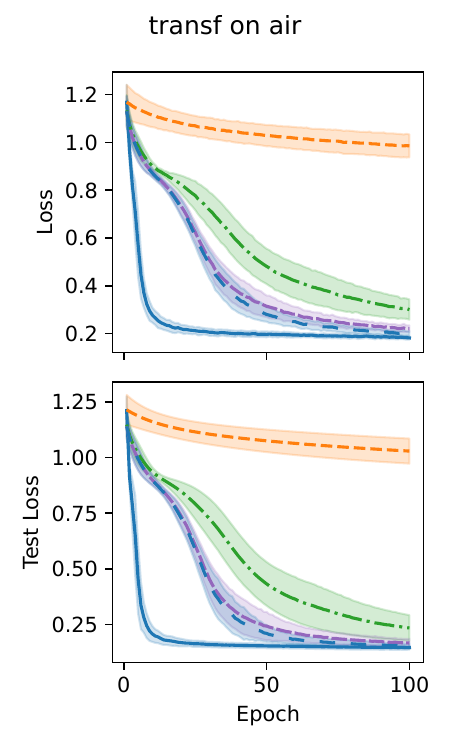}
    \caption{The effect of calibrated noisy gradient covariance as a preconditioner (blue lines) for a classification task (telescope dataset) with an MLP, a time series classification (occupancy detection dataset) and regression (air quality dataset) with a 1D CNN and a transformer encoder.}
    \label{fig:pc_cnn_transf}
\end{figure}
\paragraph{Computational cost.}
Tables~\ref{tab:epoch_time} and~\ref{tab:time_to_threshold} report wall-clock times for the optimisation experiments in Figure~\ref{fig:pc_cnn_transf}. Table~\ref{tab:epoch_time} shows per-epoch time for each method, including preconditioner calculation. The numbers in Table~\ref{tab:time_to_threshold} correspond to the wall-clock time needed to reach within 10\% of the best loss achieved in each experiment. The reported results are mean $\pm$ one standard deviation over 10 seeds. SophiaC has approximately the same per-epoch cost as Sophia (see Table~\ref{tab:epoch_time}). Despite its higher per-epoch cost, blockwise CNGC reaches the target loss fastest on all five tasks; SophiaC reaches the target loss faster than Sophia on all five tasks except \textit{Occupancy} (see Table~\ref{tab:time_to_threshold}).

\begin{table}[t]
\centering
\caption{Wall-clock time per training epoch (seconds). The results are mean $\pm$ one standard deviation over 10 random seeds.}
\label{tab:epoch_time}
\small
\setlength{\tabcolsep}{4pt}
\begin{tabular}{lccccc}
\hline \noalign{\vskip 2pt}
Optimiser &
\shortstack{MLP\\Telescope} &
\shortstack{CNN\\Occupancy} &
\shortstack{CNN\\Air} &
\shortstack{Transformer\\Occupancy} &
\shortstack{Transformer\\Air} \\
\hline
GD
& $0.22 \pm 0.00$ & $0.42 \pm 0.01$ & $0.18 \pm 0.00$
& $3.85 \pm 0.03$ & $1.81 \pm 0.02$ \\
Adam
& $0.25 \pm 0.00$ & $0.46 \pm 0.01$ & $0.20 \pm 0.00$
& $4.05 \pm 0.06$ & $1.89 \pm 0.02$ \\
Sophia
& $0.26 \pm 0.00$ & $0.49 \pm 0.01$ & $0.22 \pm 0.00$
& $4.42 \pm 0.06$ & $2.05 \pm 0.02$ \\
SophiaC (ours)
& $0.26 \pm 0.00$ & $0.50 \pm 0.00$ & $0.22 \pm 0.00$
& $4.41 \pm 0.04$ & $2.06 \pm 0.02$ \\
GD, PC (ours)
& $0.35 \pm 0.01$ & $0.66 \pm 0.01$ & $0.29 \pm 0.00$
& $4.90 \pm 0.05$ & $2.29 \pm 0.02$ \\
\hline
\end{tabular}%
\end{table}

\begin{table}[t]
\centering
\caption{Wall-clock time (seconds) to reach within 10\% of the best test loss achieved in each experiment. The corresponding thresholds are listed below. $\dagger$ indicates that the threshold was not reached within the training budget; the full elapsed training time is reported instead. Bold and italic indicate the fastest and second-fastest methods, respectively. The results are mean $\pm$ one standard deviation over 10 random seeds.}
\label{tab:time_to_threshold}
\small
\setlength{\tabcolsep}{4pt}
\begin{tabular}{lccccc}
\hline \noalign{\vskip 2pt}
Optimiser &
\shortstack{MLP\\Telescope} &
\shortstack{CNN\\Occupancy} &
\shortstack{CNN\\Air} &
\shortstack{Transformer\\Occupancy} &
\shortstack{Transformer\\Air} \\
\hline
Test-loss threshold
& $0.3402$ & $0.1701$ & $0.4804$ & $0.1767$ & $0.2526$ \\
\hline
GD
& $22.29 \pm 0.23^{\dagger}$ & $8.36 \pm 0.11^{\dagger}$
& $18.22 \pm 0.35^{\dagger}$ & $77.08 \pm 0.66^{\dagger}$
& $181.03 \pm 2.00^{\dagger}$ \\
Adam
& $16.02 \pm 0.17$ & $\mathit{3.81 \pm 0.05}$
& $20.34 \pm 0.22^{\dagger}$ & $44.71 \pm 0.66$
& $169.94 \pm 2.04$ \\
Sophia
& $7.62 \pm 0.09$ & $4.10 \pm 0.04$ & $16.78 \pm 0.28$
& $44.37 \pm 0.56$ & $96.71 \pm 0.76$ \\
SophiaC (ours)
& $\mathit{7.42 \pm 0.11}$ & $4.12 \pm 0.05$
& $\mathit{12.00 \pm 0.14}$ & $\mathit{39.90 \pm 0.38}$
& $\mathit{86.46 \pm 0.92}$ \\
GD, PC (ours)
& $\mathbf{1.65 \pm 0.02}$ & $\mathbf{0.96 \pm 0.02}$
& $\mathbf{2.60 \pm 0.02}$ & $\mathbf{10.01 \pm 0.11}$
& $\mathbf{16.32 \pm 0.16}$ \\
\hline
\end{tabular}%
\end{table}

\subsection{CNGC estimator for an early stopping criterion}\label{sec:early_stopping}
\begin{figure}
    \centering
    \includegraphics[width=0.205\linewidth,trim={10 0 10 0},clip]{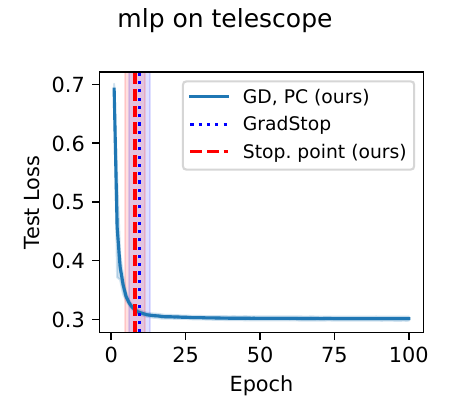}
    \includegraphics[width=0.19\linewidth,trim={25 0 10 0},clip]{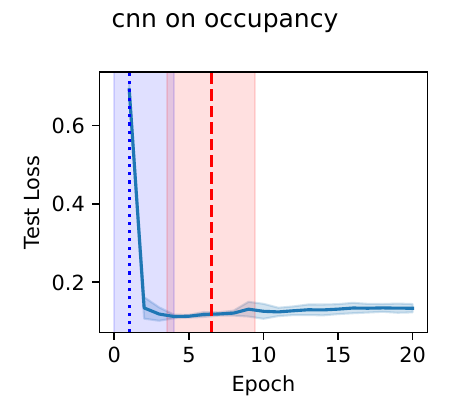}
    \includegraphics[width=0.19\linewidth,trim={25 0 10 0},clip]{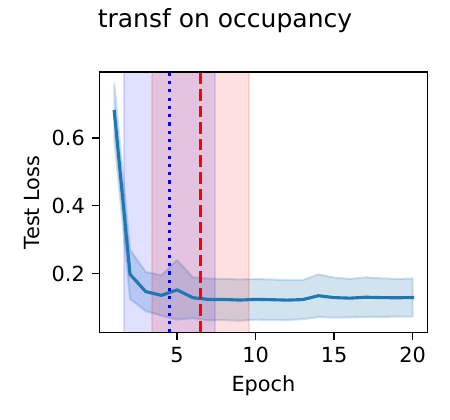}
    \includegraphics[width=0.19\linewidth,trim={25 0 10 0},clip]{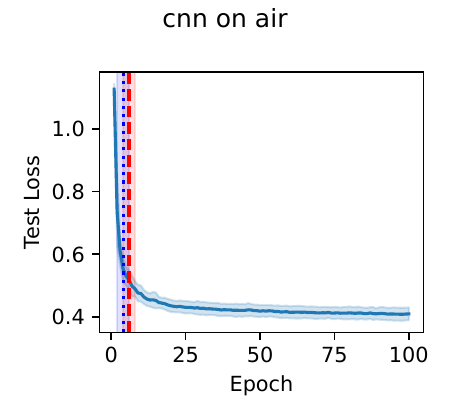}
    \includegraphics[width=0.19\linewidth,trim={25.5 0 10 0},clip]{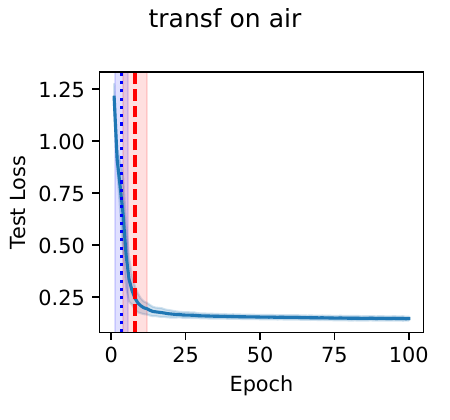}
    \caption{The effect of calibrated noisy gradient covariance for early stopping (red vertical lines) for the same tasks as in the optimisation experiment compared to the baseline.}
    \label{fig:early_stop}
\end{figure}

In this section, we demonstrate that the proposed GGN estimator can improve an existing criterion for early stopping. We use the following stopping criterion $z = \bar g^T(w)S_g^{-1}(w) \bar g(w)$, where $\bar g(w)$ is the mean of the batch gradients and $S_g^{-1}(w)$ is the inverse of the gradient covariance matrix. The training should be stopped when the value of the statistic $z$ increases. A version of this statistic was proposed in the GradStop method \citep{jamshidi2025}; however, GradStop uses a covariance matrix of the clean gradients and calculates the full covariance matrix from the gradients of the individual data points. To make it scalable, we used only the diagonal blocks of the covariance matrix and constructed it from the mean batch gradients. Thus, the baseline used here is a scalable approximation to the original GradStop construction rather than exactly the same estimator. We compared GradStop with the stopping criterion where $S_g(w)$ is the GGN estimate obtained by the proposed CNGC method (the same as the preconditioner in the previous section). The CNGC is a principled curvature estimator which leads to suitable stopping points (see Figure~\ref{fig:early_stop}), while the baseline may stop prematurely in some cases (see Figure~\ref{fig:early_stop_ind}), resulting in underfitting.

\section{Conclusion and future work}

We introduced the CNGC estimator, a gradient-only method for curvature recovery from aggregate mini-batch gradients. By injecting target noise calibrated to the local output curvature, CNGC recovers the GGN matrix in expectation and, for feed-forward ReLU networks, the diagonal Hessian blocks. Our empirical results demonstrate that CNGC accurately estimates curvature, compares favourably with baseline estimators, improves optimisation, and provides a signal for early stopping.

An interesting direction for future work is to explore the broader potential of calibrated target noise as a general-purpose probing mechanism for extracting geometric information about the loss landscape. Beyond recovering the GGN matrix, this approach could be extended to estimate other curvature-related quantities. Moreover, the recovered GGN could support a wider range of downstream applications, including uncertainty quantification, curvature-aware optimization, and adaptive regularization.

\textbf{Limitations}
Our method targets the GGN matrix rather than the full Hessian except in special cases, such as GLMs and the diagonal blocks of feed-forward ReLU networks.
The centred estimator also requires an additional backward pass with clean targets, thereby increasing computational cost.
Finally, the method assumes access to per-example model outputs and targets within each minibatch. In practice, we use diagonal or block-diagonal approximations of GGN to keep matrix inversion and storage tractable ($O(d^3)$ for inversion, $O(d^2)$ per block for storage, where $d$ is the block size).

\textbf{Societal impact} 
The paper is foundational research and is not tied to any specific applications. In practice, the results can be used, e.g., to build a more efficient optimiser for deep learning applications.

\subsection*{Acknowledgements and Disclosures}

The authors declare no competing interests that could create a conflict of interest. 
We thank the Research Council of Finland for funding, grants 364226 (Virtual Laboratory for Molecular Level Atmospheric Transformations (VILMA) Centre of Excellence) and 345704 (Finnish Center for Artificial Intelligence (FCAI)). We also thank the Helsinki Institute for Information Technology HIIT for funding. We thank the Finnish Computing Competence Infrastructure (FCCI) and CSC -- IT Center for Science, Finland -- for computational resources. The authors acknowledge the research environment provided by ELLIS Institute Finland.
Aristides Gionis
was supported by the Swedish Research Council project ExCLUS (2024-05603), the Marie Curie
Doctoral Training Network ARMADA (101168951), and the Wallenberg AI, Autonomous Systems
and Software Program (WASP) funded by the Knut and Alice Wallenberg Foundation.

\newpage

\bibliographystyle{plainnat}
\bibliography{neurips2026-ref}

\newpage


\appendix

\section{Generalized Linear Models}\label{prelem:glm}

Generalised linear models (GLMs) extend linear regression by allowing the response variable to belong to the exponential family, while preserving a linear dependence on the covariates through the natural parameter. In the notation of this paper, a GLM is the special case of a scalar-output model
\begin{equation}
    f_w(x)=w^\top x,
\end{equation}
with loss
\begin{equation}
    \ell(f_w(x),y)=A(f_w(x))-y\,f_w(x).
\end{equation}
For a batch \(B=\{(x_i,y_i)\}_{i=1}^n\), the empirical loss is therefore
\begin{equation}
    \frac{1}{n}\sum_{i=1}^n \big(A(w^\top x_i)-y_i\,w^\top x_i\big).
\end{equation}

The corresponding batch gradient is
\begin{equation}
    g^{(B)}(w)
    =
    \frac{1}{n}\sum_{i=1}^n \big(A'(w^\top x_i)-y_i\big)x_i,
\end{equation}
and the empirical Hessian is
\begin{equation}
    H^{(B)}(w)
    =
    \frac{1}{n}\sum_{i=1}^n A''(w^\top x_i)\,x_i x_i^\top.
\end{equation}
The population Hessian is
\begin{equation}
    H(w)
    =
    \mathbb{E}_{(x,y)\sim\mathcal D}
    \big[A''(w^\top x)\,x x^\top\big].
\end{equation}

Thus, in a GLM, the curvature is determined by the second derivative \(A''(\cdot)\). In particular, since \(J_w(x)=\nabla_w f_w(x)=x\), the GGN defined in Section~\ref{sec:tech} coincides exactly with the Hessian.

We briefly recall two standard examples.

\paragraph{Linear regression.}
For linear regression with squared loss,
\begin{equation}
    A(\eta)=\frac{1}{2}\eta^2.
\end{equation}
Hence,
\begin{equation}
    A'(\eta)=\eta,
    \qquad
    A''(\eta)=1.
\end{equation}
Therefore,
\begin{equation}
    H^{(B)}(w)=\frac{1}{n}\sum_{i=1}^n x_i x_i^\top,
\end{equation}
which is simply the empirical covariance matrix of the inputs.

\paragraph{Logistic regression.}
For binary logistic regression,
\begin{equation}
    A(\eta)=\log(1+e^\eta).
\end{equation}
Its derivatives are
\begin{equation}
    A'(\eta)=\sigma(\eta),
    \qquad
    A''(\eta)=\sigma(\eta)\bigl(1-\sigma(\eta)\bigr),
\end{equation}
where
\begin{equation}
    \sigma(\eta)=\frac{1}{1+e^{-\eta}}
\end{equation}
is the logistic sigmoid function. In this case, the curvature depends on \(w^\top x\) and satisfies
\begin{equation}
    0<A''(\eta)\le \frac{1}{4}.
\end{equation}

These examples illustrate the main difference between linear regression and general GLMs: in linear regression the curvature is constant, whereas in general GLMs it depends on both the current parameter and the input through \(A''(w^\top x)\).

\section{Proofs}

\subsection{Proof of Theorem~\ref{thm:gradcovGGN}}\label{sec:GGNproof}

Define
\begin{equation}
    s_i(w)
    :=
    \big(A'(f_w(x_i))-y_i\big)\,J_w(x_i).
\end{equation}
Then
\begin{equation}
    g^{(B)}(w)
    =
    \frac{1}{n}\sum_{i=1}^n s_i(w),
\end{equation}
and, by definition of \(\tilde g^{(B)}(w)\),
\begin{equation}
    \tilde g^{(B)}(w)
    =
    \frac{1}{n}\sum_{i=1}^n s_i(w)
    -
    \frac{\sqrt{\beta}}{n}\sum_{i=1}^n \xi_i J_w(x_i).
\end{equation}
Hence
\begin{equation}
    \tilde g^{(B)}(w)-g^{(B)}(w)
    =
    -\frac{\sqrt{\beta}}{n}\sum_{i=1}^n \xi_i J_w(x_i).
\end{equation}

By assumption, we have
\begin{equation}
    \mathbb{E}[\xi_i\mid x_i]=0,
    \qquad
    \mathbb{E}[\xi_i^2\mid x_i]=n\,A''(f_w(x_i)).
\end{equation}

We first prove the identity for the centred estimator. Here, we condition on the batch \(B\), so the only remaining randomness is due to the injected noise \(\xi\). Since
\[
    \widehat S_{\mathrm{cent}}^{(B)}(w)
    =
    \frac{1}{\beta}
    \bigl(\tilde g^{(B)}(w)-g^{(B)}(w)\bigr)
    \bigl(\tilde g^{(B)}(w)-g^{(B)}(w)\bigr)^\top,
\]
it suffices to show that the conditional covariance of \(\tilde g^{(B)}(w)-g^{(B)}(w)\) is \(\beta\,G^{(B)}(w)\). Since
\begin{equation}
    \mathbb{E}_{\xi}\big[\tilde g^{(B)}(w)-g^{(B)}(w)\mid B\big]=0,
\end{equation}
we have
\begin{equation}
    \mathrm{Cov}_{\xi}\big(\tilde g^{(B)}(w)-g^{(B)}(w)\mid B\big)
    =
    \frac{\beta}{n^2}
    \mathrm{Cov}_{\xi}\Big(\sum_{i=1}^n \xi_i J_w(x_i)\,\Big|\,B\Big).
\end{equation}
By independence across samples,
\begin{equation}
    \mathrm{Cov}_{\xi}\big(\tilde g^{(B)}(w)-g^{(B)}(w)\mid B\big)
    =
    \frac{\beta}{n^2} \sum_{i=1}^n
    \mathrm{Cov}_{\xi}\big(\xi_i J_w(x_i)\mid B\big).
\end{equation}
Since \(\mathbb{E}_{\xi}[\xi_i J_w(x_i)\mid B]=0\),
\begin{equation}
    \mathrm{Cov}_{\xi}\big(\xi_i J_w(x_i)\mid B\big)
    =
    \mathbb{E}_{\xi}\Big[
        \xi_i^2
        J_w(x_i)J_w(x_i)^\top
        \,\Big|\,B
    \Big].
\end{equation}
Using \(\mathbb{E}_{\xi}[\xi_i^2\mid x_i]=n\,A''(f_w(x_i))\), we obtain
\begin{align}
    \mathbb{E}_{\xi}\Big[
        \xi_i^2
        J_w(x_i)J_w(x_i)^\top
        \,\Big|\,B
    \Big]
    &=
    n\,A''(f_w(x_i))\,J_w(x_i)J_w(x_i)^\top.
\end{align}
Substituting back gives
\begin{align}
    \mathrm{Cov}_{\xi}\big(\tilde g^{(B)}(w)-g^{(B)}(w)\mid B\big)
    &=
    \frac{\beta}{n^2}\sum_{i=1}^n
    n\,A''(f_w(x_i))\,J_w(x_i)J_w(x_i)^\top \\
    &=
    \beta\,\frac{1}{n}\sum_{i=1}^n
    A''(f_w(x_i))\,J_w(x_i)J_w(x_i)^\top \\
    &=
    \beta\,G^{(B)}(w).
\end{align}
Therefore,
\begin{equation}
    \mathbb{E}_{\xi}\big[\widehat S_{\mathrm{cent}}^{(B)}(w)\mid B\big]
    =
    \frac{1}{\beta}\,
    \mathrm{Cov}_{\xi}\big(\tilde g^{(B)}(w)-g^{(B)}(w)\mid B\big)
    =
    G^{(B)}(w).
\end{equation}

We now prove the identity for the swamping estimator. In this case, we take covariance over all sources of randomness: the injected noise \(\xi\) and the random batch \((x_i,y_i)_{i=1}^n\). First,
\begin{equation}
    \mathbb{E}_{\xi,x,y}\big[\tilde g^{(B)}(w)\big]
    =
    \mathbb{E}_{x,y}\big[g^{(B)}(w)\big],
\end{equation}
so the perturbation does not bias the mean gradient. By independence across samples,
\begin{equation}
    \mathrm{Cov}_{\xi,x,y}\big(\tilde g^{(B)}(w)\big)
    =
    \frac{1}{n^2}\sum_{i=1}^n
    \mathrm{Cov}_{\xi,x,y}\big(s_i(w)-\sqrt{\beta}\,\xi_i J_w(x_i)\big).
\end{equation}
For simplicity of the notation, we do not use $\xi,x,y$ as a subscript for every $\mathrm{Cov}(\cdot)$. Since the triples \((x_i,y_i,\xi_i)\) are i.i.d., it suffices to compute one summand. For each \(i\),
\begin{align}
    \mathrm{Cov}_{\xi,x,y}\big(s_i(w)-\sqrt{\beta}\,\xi_i J_w(x_i)\big)
    &=
    \mathrm{Cov}\big(s_i(w)\big)
    +
    \beta\,\mathrm{Cov}\big(\xi_i J_w(x_i)\big) \notag\\
    &\quad
    -
    \sqrt{\beta}\,\mathrm{Cov}\big(s_i(w),\xi_i J_w(x_i)\big)
    -
    \sqrt{\beta}\,\mathrm{Cov}\big(\xi_i J_w(x_i),s_i(w)\big).
\end{align}

The cross terms vanish. Indeed,
\begin{align}
    \mathbb{E}\big[s_i(w)(\xi_i J_w(x_i))^\top\big]
    &=
    \mathbb{E}\Big[
        s_i(w)\,
        \mathbb{E}[\xi_i\mid x_i,y_i]\,
        J_w(x_i)^\top
    \Big]
    =0,
\end{align}
since \(\mathbb{E}[\xi_i\mid x_i,y_i]=\mathbb{E}[\xi_i\mid x_i]=0\). Similarly,
\begin{equation}
    \mathbb{E}\big[(\xi_i J_w(x_i))s_i(w)^\top\big]=0,
\end{equation}
and
\begin{equation}
    \mathbb{E}\big[\xi_i J_w(x_i)\big]
    =
    \mathbb{E}\Big[
        \mathbb{E}[\xi_i\mid x_i]\,J_w(x_i)
    \Big]
    =0.
\end{equation}
Thus,
\begin{equation}
    \mathrm{Cov}\big(s_i(w),\xi_i J_w(x_i)\big)=0,
    \qquad
    \mathrm{Cov}\big(\xi_i J_w(x_i),s_i(w)\big)=0.
\end{equation}

Moreover,
\begin{equation}
    \mathrm{Cov}\big(\xi_i J_w(x_i)\big)
    =
    \mathbb{E}\Big[
        \xi_i^2 J_w(x_i)J_w(x_i)^\top
    \Big]
    =
    n\,G(w),
\end{equation}
where we used the tower property exactly as above and then averaged over the batch randomness. Therefore,
\begin{equation}
    \beta\,\mathrm{Cov}\big(\xi_i J_w(x_i)\big)
    =
    \beta n\,G(w).
\end{equation}

Now define
\begin{equation}
    \Delta(w)
    :=
    \mathrm{Cov}\Big(
        \big(A'(f_w(x))-y\big)\,J_w(x)
    \Big).
\end{equation}
This quantity depends only on the data distribution and the parameter \(w\), and is therefore independent of the distribution of the added target noise. Since \(\mathrm{Cov}(s_i(w))=\Delta(w)\), we obtain
\begin{equation}
    \mathrm{Cov}\big(s_i(w)-\sqrt{\beta}\,\xi_i J_w(x_i)\big)
    =
    \Delta(w)+\beta n\,G(w).
\end{equation}
Substituting back gives
\begin{align}
    \mathrm{Cov}_{\xi,x,y}\big(\tilde g^{(B)}(w)\big)
    &=
    \frac{1}{n^2}\sum_{i=1}^n
    \Big(
        \Delta(w)+\beta n\,G(w)
    \Big) \\
    &=
    \frac{1}{n}\,\Delta(w)+\beta\,G(w).
\end{align}
Dividing by \(\beta\) yields
\begin{equation}
    \frac{1}{\beta}\,
    \mathrm{Cov}_{\xi,x,y}\big(\tilde g^{(B)}(w)\big)
    =
    G(w)+\frac{1}{n\beta}\,\Delta(w).
\end{equation}

Finally, by the moment assumptions,
\begin{equation}
    \mathbb{E}\Big[
        \big(A'(f_w(x))-y\big)^2
        \|J_w(x)\|_2^2
    \Big]
    < \infty,
\end{equation}
which implies
\begin{equation}
    \|\Delta(w)\|_{\mathrm{op}}<\infty.
\end{equation}
Hence,
\begin{equation}
    \left\|
        \frac{\Delta(w)}{n\beta}
    \right\|_{\mathrm{op}}
    =
    \frac{1}{n\beta}\,\|\Delta(w)\|_{\mathrm{op}}
    =
    \mathcal{O}\!\left(\frac{1}{n\beta}\right).
\end{equation}
This proves the theorem.

\subsection{Proof of Proposition~\ref{prop:relu_block_hessian}}

Fix an input \(x\) and a parameter \(w\) such that all pre-activations are non-zero. Then there exists a neighbourhood of \(w\) in which the ReLU masks remain unchanged. In that neighbourhood, the network can be written as
\begin{equation}
    f_{w'}(x)
    =
    W_L'
    D_{L-1}(x)
    W_{L-1}'
    \cdots
    D_1(x)
    W_1' x,
\end{equation}
where each \(D_\ell(x)\) is a fixed diagonal matrix with entries in \(\{0,1\}\).

Fix a layer \(j\), keep all parameters except \(W_j\) fixed, and write
\begin{equation}
    u_j := \mathrm{vec}(W_j).
\end{equation}
Since the network is linear in \(W_j\) when all other layers are fixed, there exists a radius \(r_j(x,w)>0\) and a vector
\begin{equation}
    \phi_j(x;w)\in\mathbb{R}^{\dim(u_j)}
\end{equation}
such that for every \(u_j'\) with
\begin{equation}
    \|u_j'-u_j\|_2<r_j(x,w),
\end{equation}
one has
\begin{equation}
    f_{(u_j',w_{-j})}(x)={u_j'}^\top \phi_j(x;w).
\end{equation}
Hence, on this neighbourhood,
\begin{equation}
    \ell(u_j';x,y)
    =
    A\big({u_j'}^\top \phi_j(x;w)\big)-y\,{u_j'}^\top \phi_j(x;w),
\end{equation}
which is exactly a linear model with GLM loss in the block variable \(u_j'\).

Therefore,
\begin{equation}
    \nabla_{u_j}\ell(w;x,y)
    =
    \big(A'(f_w(x))-y\big)\,\phi_j(x;w),
\end{equation}
and
\begin{equation}
    \nabla_{u_j}^2\ell(w;x,y)
    =
    A''(f_w(x))\,\phi_j(x;w)\phi_j(x;w)^\top.
\end{equation}
Since
\begin{equation}
    \phi_j(x;w)=\nabla_{u_j}f_w(x),
\end{equation}
it follows that
\begin{equation}
    \nabla_{u_j}^2\ell(w;x,y)
    =
    A''(f_w(x))\,
    \nabla_{u_j}f_w(x)\nabla_{u_j}f_w(x)^\top.
\end{equation}
Taking the expectation gives
\begin{equation}
    H_{jj}(w)
    =
    \mathbb{E}\Big[
        A''(f_w(x))\,
        \nabla_{u_j}f_w(x)\nabla_{u_j}f_w(x)^\top
    \Big].
\end{equation}
Similarly, for a fixed batch \(B=\{(x_i,y_i)\}_{i=1}^n\), define
\begin{equation}
    H_{jj}^{(B)}(w)
    :=
    \frac{1}{n}\sum_{i=1}^n
    A''(f_w(x_i))\,
    \nabla_{u_j}f_w(x_i)\nabla_{u_j}f_w(x_i)^\top.
\end{equation}

Now define
\begin{equation}
    s_{j,i}(w)
    :=
    \big(A'(f_w(x_i))-y_i\big)\,\nabla_{u_j}f_w(x_i),
\end{equation}
so that
\begin{equation}
    g_{u_j}^{(B)}(w)
    =
    \frac{1}{n}\sum_{i=1}^n s_{j,i}(w),
\end{equation}
and
\begin{equation}
    \tilde g_{u_j}^{(B)}(w)
    =
    \frac{1}{n}\sum_{i=1}^n s_{j,i}(w)
    -
    \frac{\sqrt{\beta}}{n}\sum_{i=1}^n \xi_i \nabla_{u_j}f_w(x_i).
\end{equation}
Hence
\begin{equation}
    \tilde g_{u_j}^{(B)}(w)-g_{u_j}^{(B)}(w)
    =
    -\frac{\sqrt{\beta}}{n}\sum_{i=1}^n \xi_i \nabla_{u_j}f_w(x_i).
\end{equation}

We first prove the identity for the centred estimator. Here, we condition on the batch \(B\), so the only remaining randomness is due to the injected noise \(\xi\). Since
\begin{equation}
    \mathbb{E}_{\xi}\big[\tilde g_{u_j}^{(B)}(w)-g_{u_j}^{(B)}(w)\mid B\big]=0,
\end{equation}
we have
\begin{equation}
    \mathrm{Cov}_{\xi}\big(\tilde g_{u_j}^{(B)}(w)-g_{u_j}^{(B)}(w)\mid B\big)
    =
    \frac{\beta}{n^2}
    \mathrm{Cov}_{\xi}\Big(
        \sum_{i=1}^n \xi_i \nabla_{u_j}f_w(x_i)
        \,\Big|\, B
    \Big).
\end{equation}
By independence across samples,
\begin{equation}
    \mathrm{Cov}_{\xi}\big(\tilde g_{u_j}^{(B)}(w)-g_{u_j}^{(B)}(w)\mid B\big)
    =
    \frac{\beta}{n^2}\sum_{i=1}^n
    \mathrm{Cov}_{\xi}\big(\xi_i \nabla_{u_j}f_w(x_i)\mid B\big).
\end{equation}
Since \(\mathbb{E}_{\xi}[\xi_i \nabla_{u_j}f_w(x_i)\mid B]=0\),
\begin{equation}
    \mathrm{Cov}_{\xi}\big(\xi_i \nabla_{u_j}f_w(x_i)\mid B\big)
    =
    \mathbb{E}_{\xi}\Big[
        \xi_i^2
        \nabla_{u_j}f_w(x_i)\nabla_{u_j}f_w(x_i)^\top
        \,\Big|\, B
    \Big].
\end{equation}
Using \(\mathbb{E}_{\xi}[\xi_i^2\mid x_i]=n\,A''(f_w(x_i))\), we obtain
\begin{align}
    \mathbb{E}_{\xi}\Big[
        \xi_i^2
        \nabla_{u_j}f_w(x_i)\nabla_{u_j}f_w(x_i)^\top
        \,\Big|\, B
    \Big]
    &=
    n\,A''(f_w(x_i))
    \nabla_{u_j}f_w(x_i)\nabla_{u_j}f_w(x_i)^\top.
\end{align}
Substituting back gives
\begin{align}
    \mathrm{Cov}_{\xi}\big(\tilde g_{u_j}^{(B)}(w)-g_{u_j}^{(B)}(w)\mid B\big)
    &=
    \frac{\beta}{n^2}\sum_{i=1}^n
    n\,A''(f_w(x_i))
    \nabla_{u_j}f_w(x_i)\nabla_{u_j}f_w(x_i)^\top \\
    &=
    \beta\,\frac{1}{n}\sum_{i=1}^n
    A''(f_w(x_i))
    \nabla_{u_j}f_w(x_i)\nabla_{u_j}f_w(x_i)^\top \\
    &=
    \beta\,H_{jj}^{(B)}(w).
\end{align}
Therefore,
\begin{equation}
    \mathbb{E}_{\xi}\big[\widehat S_{\mathrm{cent},u_j}^{(B)}(w)\mid B\big]
    =
    \frac{1}{\beta}\,
    \mathrm{Cov}_{\xi}\big(\tilde g_{u_j}^{(B)}(w)-g_{u_j}^{(B)}(w)\mid B\big)
    =
    H_{jj}^{(B)}(w).
\end{equation}

We now prove the identity for the swamping estimator. In this case, we take covariance over all sources of randomness: the injected noise \(\xi\) and the random batch \((x_i,y_i)_{i=1}^n\). By independence across samples,
\begin{equation}
    \mathrm{Cov}_{\xi,x,y}\big(\tilde g_{u_j}^{(B)}(w)\big)
    =
    \frac{1}{n^2}\sum_{i=1}^n
    \mathrm{Cov}_{\xi,x,y}\big(s_{j,i}(w)-\sqrt{\beta}\,\xi_i \nabla_{u_j}f_w(x_i)\big).
\end{equation}
For simplicity of the notation, we do not use $\xi,x,y$ as a subscript for every $\mathrm{Cov}(\cdot)$. Since the triples \((x_i,y_i,\xi_i)\) are i.i.d., it suffices to compute one summand. For each \(i\),
\begin{align}
    \mathrm{Cov}\big(s_{j,i}(w)-\sqrt{\beta}\,\xi_i \nabla_{u_j}f_w(x_i)\big)
    &=
    \mathrm{Cov}\big(s_{j,i}(w)\big)
    +
    \beta\,\mathrm{Cov}\big(\xi_i \nabla_{u_j}f_w(x_i)\big) \notag\\
    &\quad
    -
    \sqrt{\beta}\,\mathrm{Cov}\big(s_{j,i}(w),\xi_i \nabla_{u_j}f_w(x_i)\big)
    -
    \sqrt{\beta}\,\mathrm{Cov}\big(\xi_i \nabla_{u_j}f_w(x_i),s_{j,i}(w)\big).
\end{align}

The cross terms vanish because
\begin{equation}
    \mathbb{E}[\xi_i\mid x_i,y_i]=\mathbb{E}[\xi_i\mid x_i]=0,
\end{equation}
and therefore
\begin{equation}
    \mathrm{Cov}\big(s_{j,i}(w),\xi_i \nabla_{u_j}f_w(x_i)\big)=0,
    \qquad
    \mathrm{Cov}\big(\xi_i \nabla_{u_j}f_w(x_i),s_{j,i}(w)\big)=0.
\end{equation}

Moreover,
\begin{equation}
    \mathrm{Cov}\big(\xi_i \nabla_{u_j}f_w(x_i)\big)
    =
    \mathbb{E}\Big[
        \xi_i^2
        \nabla_{u_j}f_w(x_i)\nabla_{u_j}f_w(x_i)^\top
    \Big]
    =
    n\,H_{jj}(w),
\end{equation}
where we used the tower property exactly as above and then averaged over the batch randomness. Therefore,
\begin{equation}
    \beta\,\mathrm{Cov}\big(\xi_i \nabla_{u_j}f_w(x_i)\big)
    =
    \beta n\,H_{jj}(w).
\end{equation}

Now define
\begin{equation}
    \Delta_j(w)
    :=
    \mathrm{Cov}\Big(
        \big(A'(f_w(x))-y\big)\,\nabla_{u_j}f_w(x)
    \Big).
\end{equation}
This quantity depends only on the data distribution and the parameter \(w\), and is therefore independent of the distribution of the added target noise. Since \(\mathrm{Cov}(s_{j,i}(w))=\Delta_j(w)\), we obtain
\begin{equation}
    \mathrm{Cov}\big(s_{j,i}(w)-\sqrt{\beta}\,\xi_i \nabla_{u_j}f_w(x_i)\big)
    =
    \Delta_j(w)+\beta n\,H_{jj}(w).
\end{equation}
Substituting back gives
\begin{align}
    \mathrm{Cov}_{\xi,x,y}\big(\tilde g_{u_j}^{(B)}(w)\big)
    &=
    \frac{1}{n^2}\sum_{i=1}^n
    \Big(
        \Delta_j(w)+\beta n\,H_{jj}(w)
    \Big) \\
    &=
    \frac{1}{n}\,\Delta_j(w)+\beta\,H_{jj}(w).
\end{align}
Dividing by \(\beta\) yields
\begin{equation}
    \frac{1}{\beta}\,
    \mathrm{Cov}_{\xi,x,y}\big(\tilde g_{u_j}^{(B)}(w)\big)
    =
    H_{jj}(w)+\frac{1}{n\beta}\,\Delta_j(w).
\end{equation}
This proves the proposition.

\subsection{Proof of Proposition~\ref{prop:optimal_noise}}\label{sec:optimal_noise}

We keep the general multiplier \(\beta>0\) explicit throughout. In all four risk calculations below, \(\beta\) appears through a multiplicative factor in some terms, but it does not affect the perturbation law that minimises the  risk.
We prove the claim separately for the two estimators and the two notions of error.

\paragraph{Frobenius risk for the centred estimator.}
Fix a batch \(B=\{(x_i,y_i)\}_{i=1}^n\). For this batch,
\begin{equation}
\tilde g^{(B)}(w)-g^{(B)}(w)
=
-\sqrt{\beta}/n \sum\nolimits_{i=1}^n \xi_i J_w(x_i).
\end{equation}
Define
\begin{equation}
h^{(B)}(w):=\tilde g^{(B)}(w)-g^{(B)}(w).
\end{equation}
Then
\begin{equation}
\mathbb E[h^{(B)}(w)\mid B]=0,
\qquad
\operatorname{Cov}\big(h^{(B)}(w)\mid B\big)=\beta\,G^{(B)}(w),
\end{equation}
where
\begin{equation}
G^{(B)}(w)
:=
1/n \sum\nolimits_{i=1}^n
A''(f_w(x_i))\,J_w(x_i)J_w(x_i)^\top
\end{equation}
is the empirical GGN on the fixed batch.

The corresponding centred estimator is
\begin{equation}
\widehat S^{(B)}_{\mathrm{cent}}(w)
=
h^{(B)}(w)h^{(B)}(w)^\top/\beta.
\end{equation}
Since \(\mathbb E[\widehat S^{(B)}_{\mathrm{cent}}(w)\mid B]=G^{(B)}(w)\), its conditional Frobenius risk is
\begin{align}
\mathbb E\Big[
\bigl\|
\widehat S^{(B)}_{\mathrm{cent}}(w)-G^{(B)}(w)
\bigr\|_F^2
\,\Big|\,B
\Big]
&=
\mathbb E\Big[
\bigl\|
\widehat S^{(B)}_{\mathrm{cent}}(w)
\bigr\|_F^2
\,\Big|\,B
\Big]
-
\|G^{(B)}(w)\|_F^2
\\
&=
1/\beta^2\,
\mathbb E\Big[
\|h^{(B)}(w)\|_2^4
\,\Big|\,B
\Big]
-
\|G^{(B)}(w)\|_F^2,
\end{align}
because \(\|zz^\top\|_F^2=\|z\|_2^4\).

Now write
\begin{equation}
\alpha_i := -\sqrt{\beta}\,J_w(x_i)/n,
\qquad
h^{(B)}(w)=\sum\nolimits_{i=1}^n \alpha_i \xi_i.
\end{equation}
Then
\begin{equation}
\|h^{(B)}(w)\|_2^2
=
\sum\nolimits_{i=1}^n \|\alpha_i\|_2^2 \xi_i^2
+
2\sum\nolimits_{i<j}\langle \alpha_i,\alpha_j\rangle \xi_i\xi_j.
\end{equation}
Squaring and taking conditional expectation, independence and \(\mathbb E[\xi_i\mid x_i]=0\) imply that all odd terms vanish, so
\begin{align}
\mathbb E\Big[
\|h^{(B)}(w)\|_2^4
\,\Big|\,B
\Big]
&=
\sum\nolimits_{i=1}^n
\|\alpha_i\|_2^4\,\mathbb E[\xi_i^4\mid x_i]
\\
&\quad
+
2\sum\nolimits_{i<j}
\Big(
\|\alpha_i\|_2^2\|\alpha_j\|_2^2
+
2\langle \alpha_i,\alpha_j\rangle^2
\Big)
v_i v_j.
\end{align}
The second term depends only on the prescribed conditional variances \(v_i\), hence it is fixed. Therefore, the only part of the conditional Frobenius risk that depends on the perturbation law is
\begin{equation}
\sum\nolimits_{i=1}^n
\|\alpha_i\|_2^4\,\mathbb E[\xi_i^4\mid x_i].
\end{equation}
Since \(\|\alpha_i\|_2^4\) contains only the fixed batch-dependent factor \(\beta^2/n^4\), the minimiser over admissible perturbation laws is unchanged for arbitrary \(\beta>0\).
Thus, for fixed conditional variances, minimising the noise-induced Frobenius risk reduces to minimising \(\mathbb E[\xi_i^4\mid x_i]\) for each \(i\).

By Jensen's inequality,
\begin{equation}
\mathbb E[\xi_i^4\mid x_i]
=
\mathbb E[(\xi_i^2)^2\mid x_i]
\ge
\mathbb E[\xi_i^2\mid x_i]^2
=
v_i^2.
\end{equation}
Equality holds if and only if \(\xi_i^2\) is constant almost surely conditional on \(x_i\), that is,
\begin{equation}
|\xi_i|=\sqrt{v_i}
\qquad\text{almost surely.}
\end{equation}
Since \(\mathbb E[\xi_i\mid x_i]=0\), this is possible if and only if
\begin{equation}
\xi_i=
\begin{cases}
+\sqrt{v_i}, & \text{with probability }1/2,\\[3pt]
-\sqrt{v_i}, & \text{with probability }1/2.
\end{cases}
\end{equation}
Hence, the scaled Rademacher law minimises the noise-induced Frobenius risk of the centred estimator.

\paragraph{Frobenius risk for the swamping estimator.}
We now consider the swamping estimator. Define
\begin{equation}
c^{(B)}(w):=g^{(B)}(w)-\mathbb E[g^{(B)}(w)],
\end{equation}
so that
\begin{equation}
\tilde g^{(B)}(w)-\mathbb E[g^{(B)}(w)]
=
c^{(B)}(w)+h^{(B)}(w).
\end{equation}
The corresponding swamping estimator is
\begin{equation}
\widehat S^{(B)}_{\mathrm{swamp}}(w)
=
1/\beta\,
\bigl(
\tilde g^{(B)}(w)-\mathbb E[g^{(B)}(w)]
\bigr)
\bigl(
\tilde g^{(B)}(w)-\mathbb E[g^{(B)}(w)]
\bigr)^\top.
\end{equation}
Conditional on \(B\), the vector \(c^{(B)}(w)\) is deterministic and all perturbation randomness is carried only by \(h^{(B)}(w)\). Expanding gives
\begin{equation}
\widehat S^{(B)}_{\mathrm{swamp}}(w)
=
1/\beta\,
\Big(
c^{(B)}(w)c^{(B)}(w)^\top
+
c^{(B)}(w)h^{(B)}(w)^\top
+
h^{(B)}(w)c^{(B)}(w)^\top
+
h^{(B)}(w)h^{(B)}(w)^\top
\Big).
\end{equation}
Using \(\mathbb E[h^{(B)}(w)\mid B]=0\) and \(\mathbb E[h^{(B)}(w)h^{(B)}(w)^\top\mid B]=\beta G^{(B)}(w)\), write
\begin{equation}
\widehat S^{(B)}_{\mathrm{swamp}}(w)-G(w)
=
A^{(B)}(w)+R^{(B)}(w),
\end{equation}
where
\begin{equation}
A^{(B)}(w)
:=
1/\beta\,c^{(B)}(w)c^{(B)}(w)^\top
+
G^{(B)}(w)-G(w)
\end{equation}
is deterministic conditional on \(B\), and
\begin{equation}
R^{(B)}(w)
:=
1/\beta\,
\Big(
c^{(B)}(w)h^{(B)}(w)^\top
+
h^{(B)}(w)c^{(B)}(w)^\top
+
h^{(B)}(w)h^{(B)}(w)^\top
-
\mathbb E[h^{(B)}(w)h^{(B)}(w)^\top\mid B]
\Big).
\end{equation}
Since \(\mathbb E[R^{(B)}(w)\mid B]=0\),
\begin{equation}
\mathbb E\Big[
\|\widehat S^{(B)}_{\mathrm{swamp}}(w)-G(w)\|_F^2
\,\Big|\,B
\Big]
=
\|A^{(B)}(w)\|_F^2
+
\mathbb E\Big[
\|R^{(B)}(w)\|_F^2
\,\Big|\,B
\Big].
\end{equation}
Thus, the first term is a deterministic bias term, independent of the perturbation law, and only the second term depends on the choice of noise.

Expanding the second term gives
\begin{align}
\mathbb E\Big[
\|R^{(B)}(w)\|_F^2
\,\Big|\,B
\Big]
&=
1/\beta^2\,
\mathbb E\Big[
\|c^{(B)}(w)h^{(B)}(w)^\top+h^{(B)}(w)c^{(B)}(w)^\top\|_F^2
\,\Big|\,B
\Big]
\\
&\quad
+
1/\beta^2\,
\mathbb E\Big[
\|h^{(B)}(w)h^{(B)}(w)^\top-\mathbb E[h^{(B)}(w)h^{(B)}(w)^\top\mid B]\|_F^2
\,\Big|\,B
\Big]
\\
&\quad
+
2/\beta^2\,
\mathbb E\Big[
\Big\langle
c^{(B)}(w)h^{(B)}(w)^\top+h^{(B)}(w)c^{(B)}(w)^\top,\,
\\
&\qquad \qquad h^{(B)}(w)h^{(B)}(w)^\top-\mathbb E[h^{(B)}(w)h^{(B)}(w)^\top\mid B]
\Big\rangle_F
\,\Big|\,B
\Big],
\end{align}
where $\langle A , B \rangle_F := \mathrm{trace}(A^\top B)$ denotes inner product between two matrices. If, in addition, \(\mathbb E[\xi_i^3\mid x_i]=0\), then the last term vanishes because it is a sum of conditional third moments of \(h^{(B)}(w)\). The first term depends only on the conditional second moments of \(h^{(B)}(w)\), hence only on the prescribed variances \(v_i\). The second term is exactly the same fourth-moment term as in the centred case, multiplied by the same overall factor \(1/\beta^2\). Therefore, the perturbation-law dependence of the swamping Frobenius risk is unchanged for arbitrary \(\beta>0\), and the minimiser is again obtained by minimising \(\mathbb E[\xi_i^4\mid x_i]\).
Hence, the scaled Rademacher law also minimises the noise-induced Frobenius risk of the swamping estimator.

\paragraph{Directional risk for the centred estimator.}
Let \(v\in\mathbb R^d\) satisfy \(\|v\|_2=1\), and consider the directional error
\begin{equation}
v^\top\bigl(\widehat S^{(B)}_{\mathrm{cent}}(w)-G^{(B)}(w)\bigr)v.
\end{equation}
Since
\begin{equation}
\widehat S^{(B)}_{\mathrm{cent}}(w)
=
1/\beta\,h^{(B)}(w)h^{(B)}(w)^\top,
\end{equation}
we have
\begin{equation}
v^\top \widehat S^{(B)}_{\mathrm{cent}}(w) v
=
1/\beta\,(v^\top h^{(B)}(w))^2.
\end{equation}
Define
\begin{equation}
\gamma_i:=v^\top \alpha_i,
\qquad
T^{(B)}(v):=v^\top h^{(B)}(w)=\sum\nolimits_{i=1}^n \gamma_i \xi_i.
\end{equation}
Then
\begin{equation}
\mathbb E[T^{(B)}(v)^2\mid B]
=
\sum\nolimits_{i=1}^n \gamma_i^2 v_i
=
\beta\,v^\top G^{(B)}(w)v,
\end{equation}
and therefore
\begin{equation}
\mathbb E\Big[
\bigl(
v^\top(\widehat S^{(B)}_{\mathrm{cent}}(w)-G^{(B)}(w))v
\bigr)^2
\,\Big|\,B
\Big]
=
1/\beta^2\,\operatorname{Var}(T^{(B)}(v)^2\mid B).
\end{equation}
Now
\begin{equation}
\mathbb E[T^{(B)}(v)^4\mid B]
=
\sum\nolimits_{i=1}^n \gamma_i^4\,\mathbb E[\xi_i^4\mid x_i]
+
6\sum\nolimits_{i<j}\gamma_i^2\gamma_j^2 v_i v_j,
\end{equation}
so
\begin{equation}
\operatorname{Var}(T^{(B)}(v)^2\mid B)
=
\sum\nolimits_{i=1}^n
\gamma_i^4
\bigl(
\mathbb E[\xi_i^4\mid x_i]-v_i^2
\bigr)
+
4\sum\nolimits_{i<j}\gamma_i^2\gamma_j^2 v_i v_j.
\end{equation}
Thus, the only distribution-dependent part of the conditional directional risk is
\begin{equation}
\sum\nolimits_{i=1}^n \gamma_i^4\,\mathbb E[\xi_i^4\mid x_i],
\end{equation}
which is monotone in the conditional fourth moments. The prefactor \(1/\beta^2\) is again global and therefore does not change the minimiser.
Hence, exactly as in the Frobenius-risk argument, the directional risk is minimised by the scaled Rademacher law.

\paragraph{Directional risk for the swamping estimator.}
Let \(v\in\mathbb R^d\) satisfy \(\|v\|_2=1\). Define
\begin{equation}
d^{(B)}(v):=v^\top c^{(B)}(w),
\qquad
T^{(B)}(v):=v^\top h^{(B)}(w).
\end{equation}
Then
\begin{equation}
v^\top \widehat S^{(B)}_{\mathrm{swamp}}(w) v
=
1/\beta\,
\bigl(d^{(B)}(v)+T^{(B)}(v)\bigr)^2.
\end{equation}
Write
\begin{equation}
A_v^{(B)}
:=
1/\beta\,d^{(B)}(v)^2
+
v^\top G^{(B)}(w)v
-
v^\top G(w)v,
\end{equation}
which is deterministic conditional on \(B\). Since \(\mathbb E[T^{(B)}(v)^2\mid B]=\beta\,v^\top G^{(B)}(w)v\), we obtain
\begin{equation}
v^\top(\widehat S^{(B)}_{\mathrm{swamp}}(w)-G(w))v
=
A_v^{(B)}
+
1/\beta\,
\Big(
2d^{(B)}(v)T^{(B)}(v)
+
T^{(B)}(v)^2
-
\mathbb E[T^{(B)}(v)^2\mid B]
\Big).
\end{equation}
Therefore,
\begin{align}
\mathbb E\Big[
\bigl(
v^\top(\widehat S^{(B)}_{\mathrm{swamp}}(w)-G(w))v
\bigr)^2
\,\Big|\,B
\Big]
&=
(A_v^{(B)})^2
+
4(d^{(B)}(v))^2/\beta^2\,\mathbb E[T^{(B)}(v)^2\mid B]
\\
&\quad
+
1/\beta^2\,\operatorname{Var}(T^{(B)}(v)^2\mid B)
\\
&\quad
+
4d^{(B)}(v)/\beta^2\,
\mathbb E\Big[
T^{(B)}(v)\bigl(T^{(B)}(v)^2-\mathbb E[T^{(B)}(v)^2\mid B]\bigr)
\,\Big|\,B
\Big].
\end{align}
The last term is proportional to \(\mathbb E[T^{(B)}(v)^3\mid B]\), and therefore vanishes under the additional assumption \(\mathbb E[\xi_i^3\mid x_i]=0\). The first two terms depend only on second moments and are independent of the perturbation law. The only distribution-dependent part is
\begin{equation}
\operatorname{Var}(T^{(B)}(v)^2\mid B),
\end{equation}
which is exactly the same quantity analysed above for the centered estimator and is again monotone in \(\mathbb E[\xi_i^4\mid x_i]\). As before, the explicit factor \(1/\beta^2\) affects only the scale of the risk, not the perturbation law that minimises it.
Hence, under the vanishing-third-moment assumption, the scaled Rademacher law also minimises the conditional mean-squared directional error for the swamping estimator.

This completes the proof.

\section{Extension to multi-class classification}\label{app:multiclass_extension}

We now describe the extension to vector-valued outputs through the standard case of multi-class classification. Let
\begin{equation}
    f_w(x)=\big(f_{w,1}(x),\dots,f_{w,C}(x)\big)\in\mathbb{R}^C
\end{equation}
be the vector of logits, and let \(y\in\{e_1,\dots,e_C\}\subset\mathbb{R}^C\) denote the one-hot class label. We write the multi-class cross-entropy loss in the equivalent log-sum-exp form
\begin{equation}
    \ell(w;x,y)
    =
    \log\!\Big(\sum_{c=1}^C e^{f_{w,c}(x)}\Big)
    -
    y^\top f_w(x).
\end{equation}
Although this loss is often written in machine learning as \(-\sum_{c=1}^C y_c \log p_{w,c}(x)\), the above form is more convenient here because it makes the role of the log-partition function explicit and matches the notation used throughout the paper. Indeed, if
\begin{equation}
    p_w(x)
    :=
    \mathrm{softmax}(f_w(x)),
    \qquad
    p_{w,c}(x)
    =
    \frac{e^{f_{w,c}(x)}}{\sum_{r=1}^C e^{f_{w,r}(x)}},
\end{equation}
then
\begin{align}
    -\sum_{c=1}^C y_c \log p_{w,c}(x)
    &=
    -\sum_{c=1}^C y_c
    \log\!\left(
        \frac{e^{f_{w,c}(x)}}{\sum_{r=1}^C e^{f_{w,r}(x)}}
    \right) \\
    &=
    -\sum_{c=1}^C y_c
    \left(
        f_{w,c}(x)-\log\!\Big(\sum_{r=1}^C e^{f_{w,r}(x)}\Big)
    \right) \\
    &=
    -y^\top f_w(x)
    +
    \Big(\sum_{c=1}^C y_c\Big)
    \log\!\Big(\sum_{r=1}^C e^{f_{w,r}(x)}\Big).
\end{align}
Since \(y\) is one-hot, \(\sum_{c=1}^C y_c=1\), and therefore
\begin{equation}
    -\sum_{c=1}^C y_c \log p_{w,c}(x)
    =
    \log\!\Big(\sum_{c=1}^C e^{f_{w,c}(x)}\Big)
    -
    y^\top f_w(x).
\end{equation}

For each class \(c\), define
\begin{equation}
    J_{w,c}(x):=\nabla_w f_{w,c}(x).
\end{equation}
Then the one-sample gradient is
\begin{equation}
    \nabla_w \ell(w;x,y)
    =
    \sum_{c=1}^C
    \big(p_{w,c}(x)-y_c\big)\,J_{w,c}(x).
\end{equation}

To define the corresponding GGN, we first compute the loss's curvature with respect to the logits. Differentiating the softmax gives
\begin{equation}
    \frac{\partial p_{w,r}(x)}{\partial f_{w,s}(x)}
    =
    p_{w,r}(x)\big(\mathbf{1}_{r=s}-p_{w,s}(x)\big),
\end{equation}
so the Hessian of the log-sum-exp term with respect to the logits is
\begin{equation}
    M_w(x)
    :=
    \mathrm{Diag}(p_w(x)) - p_w(x)p_w(x)^\top.
\end{equation}
This is the multi-class analogue of the scalar factor \(A''(f_w(x))\) from the one-dimensional setting. Therefore, the GGN is obtained by pulling this output-space curvature back to parameter space through the classwise Jacobians, namely
\begin{equation}
    G(w)
    :=
    \mathbb{E}\Big[
        \sum_{r=1}^C\sum_{s=1}^C
        \big(M_w(x)\big)_{rs}\,
        J_{w,r}(x)J_{w,s}(x)^\top
    \Big].
\end{equation}

For a batch \(B=\{(x_i,y_i)\}_{i=1}^n\), we perturb the targets by
\begin{equation}
    \tilde y_i = y_i + \sqrt{\beta}\,\xi_i,
\end{equation}
where \(\xi_i\in\mathbb{R}^C\) are conditionally independent and satisfy
\begin{equation}
    \mathbb{E}[\xi_i\mid x_i]=0,
    \qquad
    \mathrm{Cov}(\xi_i\mid x_i)=n\,M_w(x_i).
\end{equation}
The corresponding perturbed batch gradient is
\begin{equation}
    \tilde g^{(B)}(w)
    :=
    \frac{1}{n}\sum_{i=1}^n
    \sum_{c=1}^C
    \big(p_{w,c}(x_i)-\tilde y_{i,c}\big)\,
    J_{w,c}(x_i).
\end{equation}

\begin{proposition}[Vector-output CNGC for multi-class classification]
\label{prop:vector_cngc_multiclass}
Assume an i.i.d.\ batch \(B=\{(x_i,y_i)\}_{i=1}^n\) satisfying the required moment conditions. Let \(\xi_i\in\mathbb{R}^C\) be conditionally independent perturbations satisfying
\[
\mathbb{E}[\xi_i\mid x_i]=0,
\qquad
\mathrm{Cov}(\xi_i\mid x_i)=nM_w(x_i),
\]
and let \(\tilde y_i=y_i+\sqrt{\beta}\xi_i\). Then the vector-output centred and swamping estimators satisfy
\begin{equation}
\mathbb{E}_\xi\!\left[
\widehat S^{(B)}_{\mathrm{cent}}(w)
\,\middle|\, B
\right]
=
G^{(B)}(w),
\end{equation}
and
\begin{equation}
\mathbb{E}_{\xi,x,y}\!\left[
\widehat S^{(B)}_{\mathrm{swamp}}(w)
\right]
=
G(w)+\Delta(w)/(n\beta),
\end{equation}
where \(\Delta(w)\) depends only on the data distribution \(\mathcal{D}\) and the parameter \(w\), and is independent of the injected-noise distribution. Moreover,
\[
\|\Delta(w)/(n\beta)\|_{\mathrm{op}}
=
\mathcal{O}(1/(n\beta)).
\]
\end{proposition}
\begin{proof}
Define
\begin{equation}
    s_i(w)
    :=
    \sum_{c=1}^C
    \big(p_{w,c}(x_i)-y_{i,c}\big)\,J_{w,c}(x_i).
\end{equation}
Then
\begin{equation}
    g^{(B)}(w)
    =
    \frac{1}{n}\sum_{i=1}^n s_i(w),
\end{equation}
and
\begin{equation}
    \tilde g^{(B)}(w)
    =
    \frac{1}{n}\sum_{i=1}^n s_i(w)
    -
    \frac{\sqrt{\beta}}{n}\sum_{i=1}^n
    \sum_{c=1}^C \xi_{i,c}\,J_{w,c}(x_i).
\end{equation}
Hence
\begin{equation}
    \tilde g^{(B)}(w)-g^{(B)}(w)
    =
    -\frac{\sqrt{\beta}}{n}\sum_{i=1}^n
    \sum_{c=1}^C \xi_{i,c}\,J_{w,c}(x_i).
\end{equation}

Define also
\begin{equation}
    z_i(w)
    :=
    \sum_{c=1}^C \xi_{i,c}\,J_{w,c}(x_i).
\end{equation}
By assumption,
\begin{equation}
    \mathbb{E}[\xi_i\mid x_i]=0,
    \qquad
    \mathrm{Cov}(\xi_i\mid x_i)=n\,M_w(x_i).
\end{equation}

We first prove the identity for the centred estimator. Here we condition on the batch \(B\), so the only remaining randomness is due to the injected noise \(\xi\). Since the centered estimator in Eq.~(10) includes the normalisation \(1/\beta\), it suffices to show that the conditional covariance of \(\tilde g^{(B)}(w)-g^{(B)}(w)\) is \(\beta\,G^{(B)}(w)\). Since
\begin{equation}
    \mathbb{E}_{\xi}\big[\tilde g^{(B)}(w)-g^{(B)}(w)\mid B\big]=0,
\end{equation}
we have
\begin{equation}
    \mathrm{Cov}_{\xi}\big(\tilde g^{(B)}(w)-g^{(B)}(w)\mid B\big)
    =
    \frac{\beta}{n^2}\sum_{i=1}^n
    \mathrm{Cov}_{\xi}\big(z_i(w)\mid B\big),
\end{equation}
where we used independence across samples. Now
\begin{align}
    \mathrm{Cov}_{\xi}\big(z_i(w)\mid B\big)
    &=
    \mathbb{E}_{\xi}\big[z_i(w)z_i(w)^\top\mid B\big] \\
    &=
    \sum_{r=1}^C\sum_{s=1}^C
    \mathbb{E}_{\xi}\big[\xi_{i,r}\xi_{i,s}\mid x_i\big]\,
    J_{w,r}(x_i)J_{w,s}(x_i)^\top \\
    &=
    n\sum_{r=1}^C\sum_{s=1}^C
    \big(M_w(x_i)\big)_{rs}\,
    J_{w,r}(x_i)J_{w,s}(x_i)^\top.
\end{align}
Substituting back gives
\begin{align}
    \mathrm{Cov}_{\xi}\big(\tilde g^{(B)}(w)-g^{(B)}(w)\mid B\big)
    &=
    \frac{\beta}{n^2}\sum_{i=1}^n
    n\sum_{r=1}^C\sum_{s=1}^C
    \big(M_w(x_i)\big)_{rs}\,
    J_{w,r}(x_i)J_{w,s}(x_i)^\top \\
    &=
    \beta\,
    \frac{1}{n}\sum_{i=1}^n
    \sum_{r=1}^C\sum_{s=1}^C
    \big(M_w(x_i)\big)_{rs}\,
    J_{w,r}(x_i)J_{w,s}(x_i)^\top \\
    &=
    \beta\,G^{(B)}(w).
\end{align}
Therefore, after the normalisation by \(1/\beta\) in Eq.~(10), the centered estimator has conditional expectation \(G^{(B)}(w)\).

We now prove the identity for the swamping estimator. First,
\begin{equation}
    \mathbb{E}_{\xi,x,y}\big[\tilde g^{(B)}(w)\big]
    =
    \mathbb{E}_{x,y}\big[g^{(B)}(w)\big],
\end{equation}
so the perturbation does not bias the mean gradient. By independence across samples,
\begin{equation}
    \mathrm{Cov}_{\xi,x,y}\big(\tilde g^{(B)}(w)\big)
    =
    \frac{1}{n^2}\sum_{i=1}^n
    \mathrm{Cov}_{\xi,x,y}\big(s_i(w)-\sqrt{\beta}\,z_i(w)\big).
\end{equation}
For each \(i\),
\begin{align}
    \mathrm{Cov}\big(s_i(w)-\sqrt{\beta}\,z_i(w)\big)
    &=
    \mathrm{Cov}\big(s_i(w)\big)
    +
    \beta\,\mathrm{Cov}\big(z_i(w)\big) \notag\\
    &\quad
    -
    \sqrt{\beta}\,\mathrm{Cov}\big(s_i(w),z_i(w)\big)
    -
    \sqrt{\beta}\,\mathrm{Cov}\big(z_i(w),s_i(w)\big).
\end{align}
The cross terms vanish because
\begin{equation}
    \mathbb{E}[z_i(w)\mid x_i,y_i]
    =
    \sum_{c=1}^C \mathbb{E}[\xi_{i,c}\mid x_i,y_i]\,J_{w,c}(x_i)
    =
    0.
\end{equation}
Hence
\begin{equation}
    \mathrm{Cov}\big(s_i(w),z_i(w)\big)=0,
    \qquad
    \mathrm{Cov}\big(z_i(w),s_i(w)\big)=0.
\end{equation}

It remains to compute \(\mathrm{Cov}(z_i(w))\). Since \(\mathbb{E}[z_i(w)]=0\),
\begin{align}
    \mathrm{Cov}\big(z_i(w)\big)
    &=
    \mathbb{E}\big[z_i(w)z_i(w)^\top\big] \\
    &=
    \mathbb{E}\Big[
        \sum_{r=1}^C\sum_{s=1}^C
        \xi_{i,r}\xi_{i,s}\,
        J_{w,r}(x_i)J_{w,s}(x_i)^\top
    \Big] \\
    &=
    \mathbb{E}\Big[
        \sum_{r=1}^C\sum_{s=1}^C
        \mathbb{E}[\xi_{i,r}\xi_{i,s}\mid x_i]\,
        J_{w,r}(x_i)J_{w,s}(x_i)^\top
    \Big] \\
    &=
    \mathbb{E}\Big[
        \sum_{r=1}^C\sum_{s=1}^C
        n\,(M_w(x_i))_{rs}\,
        J_{w,r}(x_i)J_{w,s}(x_i)^\top
    \Big] \\
    &=
    n\,G(w).
\end{align}
Therefore,
\begin{equation}
    \beta\,\mathrm{Cov}\big(z_i(w)\big)
    =
    \beta n\,G(w).
\end{equation}

Now let
\begin{equation}
    \Delta(w)
    :=
    \mathrm{Cov}\Big(
        \sum_{c=1}^C
        \big(p_{w,c}(x)-y_c\big)\,J_{w,c}(x)
    \Big).
\end{equation}
This quantity depends only on the data distribution and the parameter \(w\), and is therefore independent of the added target noise. Since \(\mathrm{Cov}(s_i(w))=\Delta(w)\), we obtain
\begin{equation}
    \mathrm{Cov}\big(s_i(w)-\sqrt{\beta}\,z_i(w)\big)
    =
    \Delta(w)+\beta n\,G(w).
\end{equation}
Substituting back gives
\begin{align}
    \mathrm{Cov}_{\xi,x,y}\big(\tilde g^{(B)}(w)\big)
    &=
    \frac{1}{n^2}\sum_{i=1}^n
    \Big(
        \Delta(w)+\beta n\,G(w)
    \Big) \\
    &=
    \frac{1}{n}\,\Delta(w)
    +
    \beta\,G(w).
\end{align}
Dividing by \(\beta\) yields
\begin{equation}
    \mathrm{Cov}_{\xi,x,y}\big(\tilde g^{(B)}(w)\big)/\beta
    =
    G(w)
    +
    \frac{1}{n\beta}\,\Delta(w).
\end{equation}
The operator-norm bound follows immediately from the finiteness assumptions.
\end{proof}

\textbf{Examples of admissible noise.} There are several choices of noise \(\xi_i\in\mathbb{R}^C\) with conditional mean zero and conditional covariance
\[
\mathrm{Cov}(\xi_i\mid x_i)=n\,M_w(x_i).
\]
A natural continuous choice is Gaussian noise,
\[
\xi_i \mid x_i \sim \mathcal{N}\big(0,\; n\,M_w(x_i)\big).
\]
A particularly simple discrete choice is obtained by sampling
\[
z_i \sim \mathrm{Categorical}(p_w(x_i)),
\]
where \(p_w(x_i)=\mathrm{softmax}(f_w(x_i))\), and then setting
\[
\xi_i := \sqrt{n}\,\big(z_i-p_w(x_i)\big).
\]
Since
\[
\mathbb{E}[z_i\mid x_i]=p_w(x_i),
\qquad
\mathrm{Cov}(z_i\mid x_i)
=
\mathrm{Diag}(p_w(x_i))-p_w(x_i)p_w(x_i)^\top
=
M_w(x_i),
\]
it follows that
\[
\mathbb{E}[\xi_i\mid x_i]=0,
\qquad
\mathrm{Cov}(\xi_i\mid x_i)=n\,M_w(x_i).
\]

Another admissible discrete choice is matrix-transformed Rademacher noise. Let \(\varepsilon_i\in\{-1,+1\}^C\) have independent, equally likely coordinates, drawn independently across samples and independently of the data, and set
\[
\xi_i := \sqrt{n}\,M_w(x_i)^{1/2}\varepsilon_i,
\]
where \(M_w(x_i)^{1/2}\) denotes the symmetric positive-semidefinite square root. Since \(\mathbb{E}[\varepsilon_i]=0\) and \(\mathrm{Cov}(\varepsilon_i)=I_C\), we obtain
\[
\mathbb{E}[\xi_i\mid x_i]=0,
\qquad
\mathrm{Cov}(\xi_i\mid x_i)
=
n\,M_w(x_i)^{1/2}\big(M_w(x_i)^{1/2}\big)^\top
=
n\,M_w(x_i).
\]
Thus, this construction also satisfies the required calibration conditions.

\section{Block-wise Preconditioned Optimisation}
\label{subsec:blockwise-preconditioned-optimisation}

As an example of how the proposed estimator can be used in optimisation,
Algorithm~\ref{alg:blockwise-preconditioned-centered} describes a block-wise
preconditioned gradient descent method. During each epoch, the method averages
rank-one centred curvature samples for each layer block. At the end of the epoch,
it forms a diagonal-block preconditioner using the pseudo-inverse, which is then used to
precondition gradients in the following epoch. Since approximating the inverse matrix is costly, we minimised computational overhead by inverting the preconditioner only once per epoch. For the swamping estimator, we estimate the centring $\mu$ as the average gradient over the same epoch. 

In the algorithm, we utilise an adaptation of the learning rate update rule from \citet{martens2015optimizing,botev2017practical}. Unlike in the work by \citet{botev2017practical}, we update the learning rate $\alpha_e$ and the regularisation strength parameter $\tau_e$ only once per epoch, using mid-epoch and end-of-epoch loss values to stabilise the updates. We also did not use an additional regularisation parameter $\gamma$ in our implementation.

\begin{algorithm}[t]
\caption{Block-wise preconditioned gradient descent with noisy-gradient curvature estimates (GD, PC).}
\label{alg:blockwise-preconditioned-centered}
\centering
\small
\newcommand{\I}{\hspace*{1.5em}}
\newcommand{\II}{\hspace*{3.0em}}
\newcommand{\III}{\hspace*{4.5em}}
\begin{tabular}{@{}r@{\hspace{0.6em}}p{0.90\linewidth}@{}}
\hline
\\[-1.8ex]

1:  & \textbf{Input:} initial parameter $w_0$, estimator type, learning-rate rule $\alpha(\cdot)$, layer blocks $u_j=\mathrm{vec}(W_j)$, $j\in[m]$. \\

2:  & Initialize $P_{j,0}=I$ for all layers $j\in[m]$, and set $\alpha_0=10^{-4}$, $\tau_0=1$, $\omega=0.9$. \\

3:  & \textbf{for} epoch $e=0,1,\ldots,E-1$ \textbf{do} \\

4:  & \I Initialize $\overline G_{j,e}=0$ for all layers $j\in[m]$. \\

5:  & \I \textbf{for} batch $t=0,1,\ldots,T_e-1$ \textbf{do} \\

6:  & \II Compute $g_t=\nabla_w \sum_i \ell(f_{w_t}(x_i),y_i) / n$. \\

7:  & \II Call Algorithm~\ref{alg:centered-swamping} with the chosen estimator type to obtain block samples $s_{j,t}$, $j\in[m]$. \\

8:  & \II \textbf{for} each layer $j\in[m]$ \textbf{do} \\

9:  & \III Extract $g_{u_j,t}$ from $g_t$. \\

10: & \III Update $\overline G_{j,e}\leftarrow \mathrm{Update}(\overline G_{j,e},s_{j,t})$. \\

11: & \III Compute $p_{j,t}=P_{j,e}g_{u_j,t}$. \\

12: & \II \textbf{end for} \\

13: & \II Set $p_t=(p_{1,t},\ldots,p_{m,t})$. \\

14: & \II Update $w_{t+1}=w_t-\alpha_e p_t$. \\

15: & \I \textbf{end for} \\

16: & \I \textbf{for} each layer $j\in[m]$ \textbf{do} \\

17: & \II Set $P_{j,e+1}=(\overline G_{j,e})^{-1}$. \\

18: & \I \textbf{end for} \\

19: & \I Update $(\alpha_{e+1},\tau_{e+1})$ using $L_e$, $L_{e-1}$, $\{P_{j,e+1}\}_{j=1}^m$, $g_t$, $p_t$, $\tau_e$, and $\omega$. \\

20: & \textbf{end for} \\

21: & \textbf{return} $w_T$. \\
\hline
\end{tabular}
\end{algorithm}

\section{Experimental details}

\subsection{Generated data and model weight perturbations} \label{sec:exp_details}
\paragraph{Generated data}
The generated data were obtained by first sampling a $ k$-dimensional matrix from a Gaussian distribution with zero mean and standard deviation 3. We then transformed the features using the same neural network used for testing (e.g., an MLP or a CNN). The network outputs were perturbed by adding small Gaussian noise to better mimic the real data. To obtain classification labels from network outputs, we converted the outputs to probabilities and sampled the labels from a binomial distribution.

\paragraph{Model perturbation}
To study the Hessian and its approximations away from optimal parameters, we shifted the true model parameters used for data generation by a vector with varying norm. We got the perturbation vector by calculating the sensitivity of the loss to changes in each model parameter. This allowed us to test the approximations relative to the true model parameters.

\subsection{Error bars reporting}
In all our experiments, except the illustrative example in Figure~\ref{fig:gnn_ngc_hessian}, we reported the mean and the 1-standard-deviation interval obtained from repeated runs with 10 random seeds. The study of the estimator variance w.r.t. the added noise distribution (Figure~\ref{fig:curv_est}, \textit{right}) was repeated 30 times.

\subsection{Curvature estimators: effects of centring, swamping, and noise design}\label{sec:curv_theory}

In this experiment, we used a small MLP with 105 parameters. The MLP had a hidden layer of size 5, and the input had 20 features; the dataset of size 50K was generated using the same regression model. A small network was selected for demonstration purposes due to memory limits, since we calculated the full Hessian for comparisons (in Figure~\ref{fig:curv_est_hessian}).  

For the evaluations of curvature and noise design, we shifted the true model parameters by vectors of norm ranging from 0.1 to 2.6. For the evaluation of the bias of our swamping estimator, the model parameters were shifted by a vector of norm 2.6, and the noise multiplier was set to 0.1, 1.0, 2.0, 5.0, and 10.0. For the curvature and swamping bias evaluations, the dataset and model perturbations were resampled for each random seed, whereas in the noise design evaluation, the only source of randomness across seeds was the added noise.
We used a batch size of 64.

\subsection{Curvature estimators: comparison with the baselines}\label{sec:curv_baselines}
We tested the curvature estimators for a classification task using an MLP and a regression task using a CNN. We used the gamma telescope dataset \citep{heck1998corsika,magic_gamma_telescope_159} for classification and the air quality dataset \citep{vito2008onfield} for regression evaluation. The MLP had two hidden layers of size 20 with ReLU non-linearities. The CNN had two convolutional layers with 20 output channels and a kernel size of five. Max pooling was applied after each convolutional layer, and the network had a linear output layer. We used a batch size of 128. For our method, we used a swamping variant of the estimator because the clean gradient was not readily available, unlike in the optimisation experiments. We set the noise multiplier equal to 10 for the evaluations. We also evaluated the centring variant of our estimator, and the results were very close.

For the comparison of the curvature estimators with the true Hessian matrix and its eigen spectrum (Figure~\ref{fig:est_appendix}, \textit{bottom row}), we used a batch size equal to 8 (to lower the memory requirements). In the evaluations of the curvature estimators, the model weights were identical across all estimators. The weights were set to their initial values. The full datasets were evaluated for comparisons with different dataset sizes. The initial values of the model weights and the sampled noise used to evaluate the GGN estimates were the sources of randomness across different random seeds.

\subsection{CNGC estimator as the preconditioner in optimisation}\label{sec:precond}

\textbf{Datasets}
Gamma Telescope \citep{heck1998corsika,magic_gamma_telescope_159}, Bank Marketing \citep{moro2014data}, Adult Income \citep{kohavi1996scaling}, Air Quality \citep{vito2008onfield}, Room Occupancy Detection \citep{candanedo2016accurate} were accessed through OpenML or UCI Machine Learning Repository.

The air quality and room occupancy detection datasets contain time-series data and were modelled using 1D CNNs and transformer encoders. Each model input consisted of 50 consecutive time steps.

\textbf{Models} \textit{MLPs} had two hidden layers of size 20, each with ReLU. \textit{CNNs} had two convolutional layers with five output channels, a kernel size of five and max pooling followed by a linear output layer. \textit{Transformer encoders} had two blocks with two attention heads each, an embedding size of 8, 8 features in the feed-forward layers, and a dropout probability of 0.05.

We used a batch size of 128 and a learning rate of $10^{-4}$ for all optimisers. The data splits were consistent across random seeds; only the initial values for the model weights and the sampled noise for the GGN estimators differed between random seeds. For the optimiser with the block-diagonal preconditioner, we accumulated gradients for the covariance matrix estimator during the full epoch. We then inverted the blocks and used them as preconditioners for parameter updates during the next epoch. Thus, the preconditioner inversion was called only once per epoch. Additionally, the rank of the matrix depends on the number of gradients used to construct the preconditioner; accumulating gradients over the entire epoch helps ensure a sufficient rank for the estimated blocks.

\textbf{Implementation} We used a PyTorch implementation of (stochastic) gradient descent (GD), Muon and Adam optimisers. To evaluate Sophia Optimiser, we used the original implementation provided by the method's authors. In our implementation of SophiaC with our preconditioner, we replace the curvature estimator with ours. The implementation of the block-diagonal preconditioner builds on top of the GD implementation and is described in detail in Algorithm~\ref{alg:blockwise-preconditioned-centered}.

\section{Additional experiments}

\subsection{Illustrative examples for GGN recovery via Calibrated noisy gradient covariance}\label{sec:illustrative}
Here, we provide illustrative examples of the true Hessian, the Generalised Gauss-Newton matrix (GGN), and the Calibrated noisy gradient covariance matrix (CNGC). The example was built with generated data with eight input features for regression and classification tasks. As the model, we used a CNN with one convolution layer, max pooling, ReLU non-linearity, followed by a linear layer and a sigmoid activation for classification. The model had 12 parameters in total. The model weights were perturbed by a vector of norm 1.1 (see Section~\ref{sec:exp_details} for details). In this experiment, the Hessian blocks coincided with the GGN blocks (because the model was a CNN with ReLU activations).

GGN recovered the diagonal blocks of the true Hessian, but its calculation required a Jacobian matrix for each data point. CNGC matrix also provided a close approximation of the true Hessian. Even though the diagonal blocks were not captured exactly, CNGC required only the batch-mean gradient for evaluation. Both GGN and CNGC are positive semi-definite by construction, and both captured the spectrum of the true Hessian well. In this example, the true Hessian was not positive definite, so we compared only the positive eigenvalues of the Hessian with those of both approximations. The relative operator norm of the difference between eigenvalues of the true Hessian and GGN and CNGC was 0.29 and 0.26 for the regression task and 0.77 and 0.75 for the classification task, respectively.
The results are illustrated below in Figure~\ref{fig:gnn_ngc_hessian}.

\begin{figure}
    \centering
    \includegraphics[width=\linewidth]{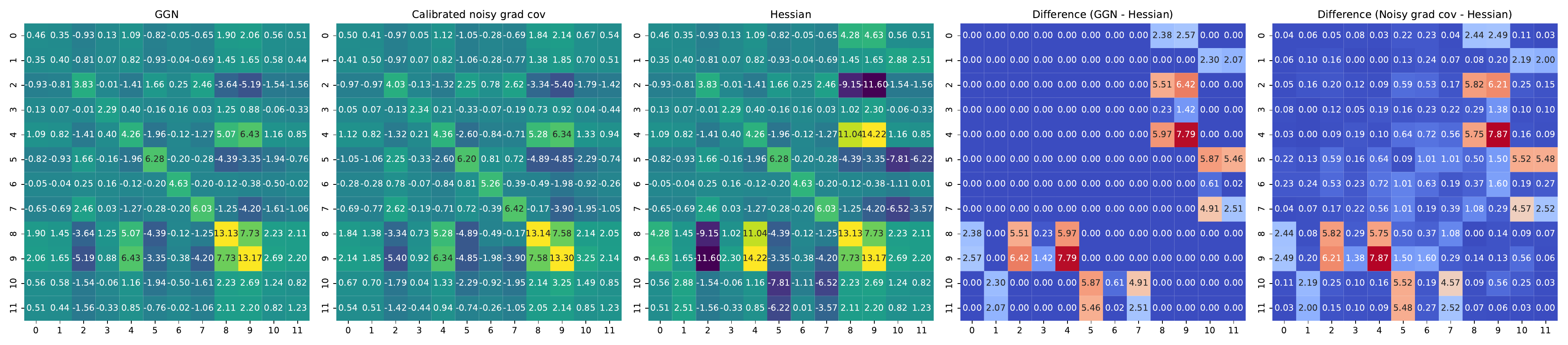}
    \includegraphics[width=\linewidth]{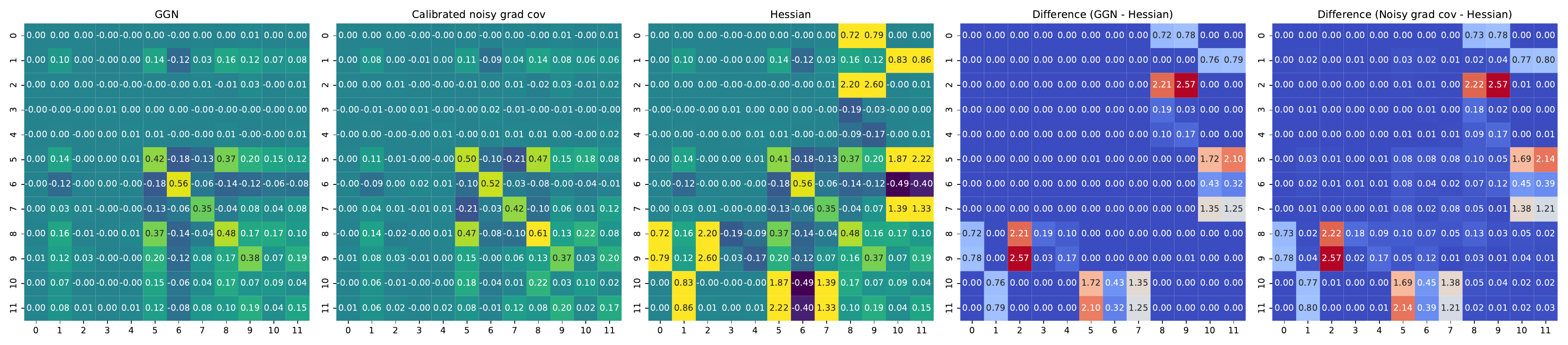}
    \caption{Illustrative example comparing CNGC and GGN with the true Hessian for a small convolutional network with 12 parameters. \textbf{Top:} regression, \textbf{bottom:} classification.}
    \label{fig:gnn_ngc_hessian}
\end{figure}

\subsection{Recovery of Hessian eigenvalues} \label{sec:h_recovery_exp}

\begin{figure}
    \centering
    \includegraphics[width=0.76\textwidth, trim={0 7 0 0}, clip]{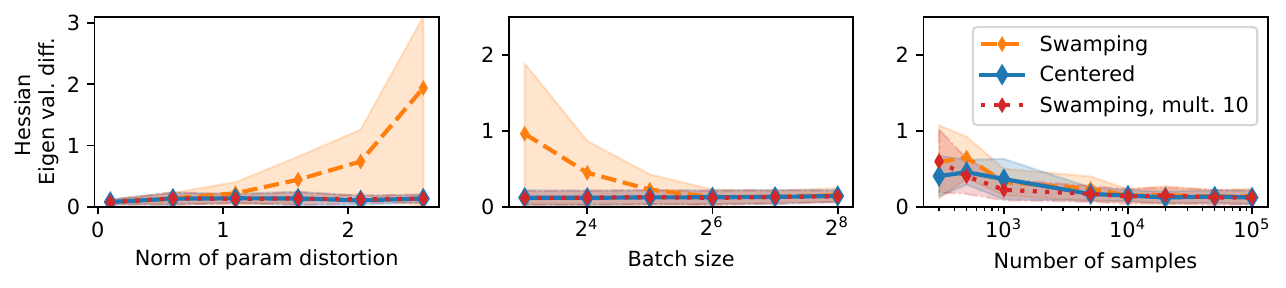}
    \caption{Relative operator norm between the spectrum of the true Hessian and the curvature estimator with centring and swamping with noise multiplier equal to 1 and to 10.  Both centring and swamping with noise, with a multiplier equal to 10, capture the eigenspectrum of the true Hessian. }
    \label{fig:curv_est_hessian}
 \end{figure}

We further tested the effects of centring and swamping on the recovery of (positive) eigenvalues of the true Hessian for various parameter perturbations, batch sizes, and the number of samples $N=Kn$ (see Figure~\ref{fig:curv_est_hessian}). Both centring and swamping with a noise multiplier equal to 10 captured the eigenspectrum of the Hessian. Both methods significantly improved the robustness of the estimate for small batch sizes and model parameters far from the optimum (\textit{left and middle}). All estimators required a sufficient number of samples in the dataset (\textit{right}); however, even for small datasets with fewer than 1000 samples, the estimates captured the true structure, especially the centred estimate. 

\subsection{Comparison of GGN estimators}
\begin{figure}
    \centering 

    \includegraphics[width=0.245\linewidth,trim={3 0 2 0},clip]{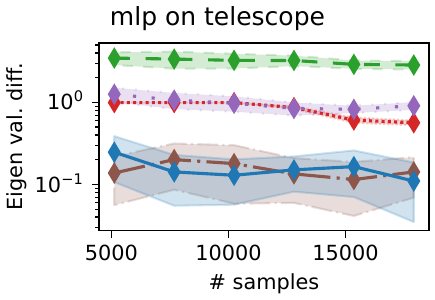}
    \includegraphics[width=0.235\linewidth,trim={13 0 2 0},clip]{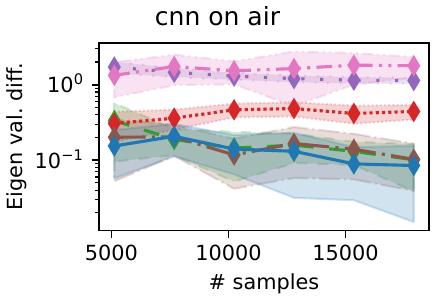}
    \includegraphics[width=0.34\linewidth,trim={3 0 2 0},clip]{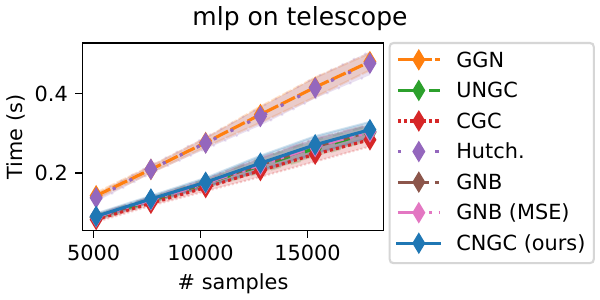}

    \includegraphics[width=0.275\linewidth,trim={3 0 2 0},clip]{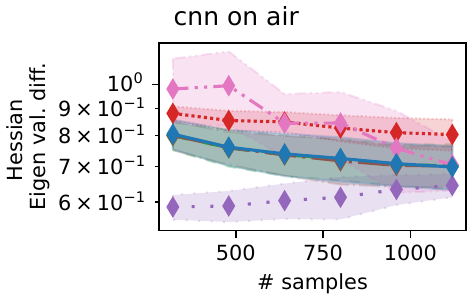}
    \includegraphics[width=0.255\linewidth,trim={3 0 2 0},clip]{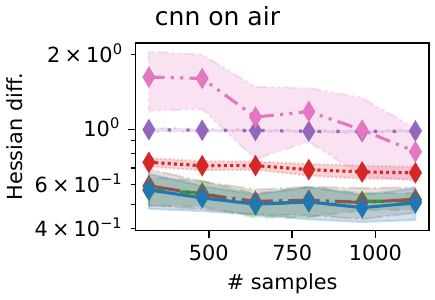}
    
    \caption{\textit{Top: }Comparison of GGN estimators with an MLP for a classification task (\textit{left}) and a CNN for a regression task (\textit{middle}) by the relative operator norm of the difference of the eigenvalue spectrum of the full GGN matrix and its estimates. The wall-clock time for the calculations of GGN estimates for the MLP model (\textit{right}). \textit{Bottom: } Comparison of GGN estimators with the true Hessian for a CNN on a regression task. The relative operator norm of the difference between (positive) Hessian eigenvalues and the eigenvalues of the estimates (\textit{left}), the relative operator norm of the difference between full matrices of the estimates and the Hessian (\textit{right}). }
    \label{fig:est_appendix}
\end{figure}

Additional comparisons of the curvature estimators for the recovery of the eigenvalue spectrum of GGN are presented in Figure~\ref{fig:est_appendix}, \textit{top} row. In this section, the experimental setup was the same as in Section~\ref{sec:curv_est_baselines}. The observations on the recovery of the full GGN matrix presented in the main text also hold for its eigenspectrum.

Furthermore, we compared the obtained GGN estimates with the eigenvalues of the Hessian and the full Hessian matrix (see Figure~\ref{fig:est_appendix}, \textit{bottom} row). Hutchinson estimator captured the eigenvalues of the true Hessian best, while both calibrated GNB and the proposed CNGC provided good estimates for the eigenvalues and recovered the full matrix best.

\subsection{CNGC estimator as the preconditioner in optimisation}\label{sec:preconditioner_app}

This section contains additional runs for the optimisation algorithms on classification tasks with logistic regression and MLP classifiers. The results for Muon optimiser are available only for those models, since the current PyTorch implementation supports only 2D weight matrices.
All first-order methods in this experiment were run with a learning rate of 1e-4. Tuning the learning rate and its schedule for each optimiser may speed up convergence, except for gradient descent, which encounters numerical difficulties at higher learning rates. The optimiser with block-diagonal preconditioner is also trained with the learning rate equal to 1e-4 during the first epoch. However the block-diagonal preconditioner enables the estimation of the optimal learning rate at each step based on the estimated GGN matrix. Thus, the learning rate was automatically adjusted for later steps. Overall, this evaluation confirms that the proposed estimator is efficient as the preconditioner for both the diagonal and the block-wise versions of the algorithms.

\begin{figure}
    \centering 
    \includegraphics[width=0.204\linewidth,trim={11 0 10 0},clip]{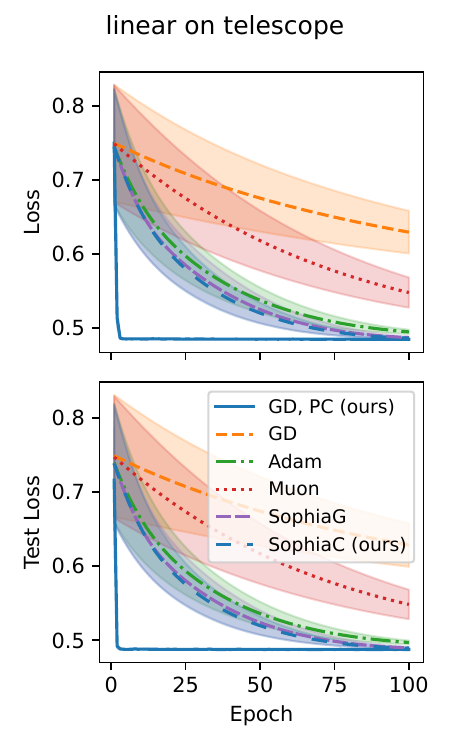}
    \includegraphics[width=0.19\linewidth,trim={25 0 10 0},clip]{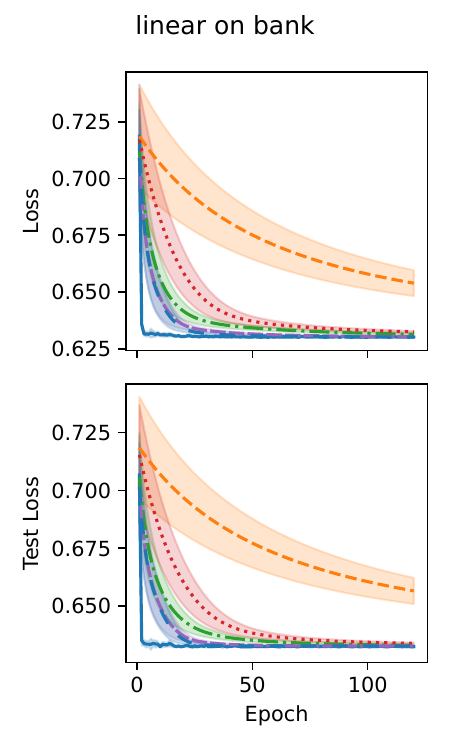}
    \includegraphics[width=0.19\linewidth,trim={25 0 10 0},clip]{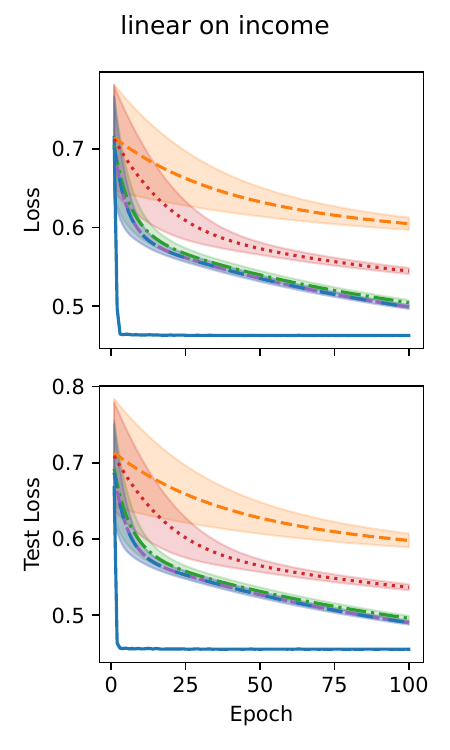}
    \includegraphics[width=0.19\linewidth,trim={25 0 10 0},clip]{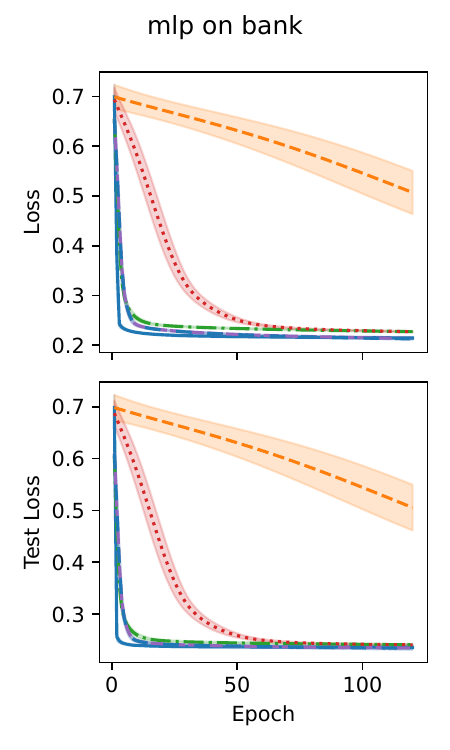}
    \includegraphics[width=0.19\linewidth,trim={25 0 10 0},clip]{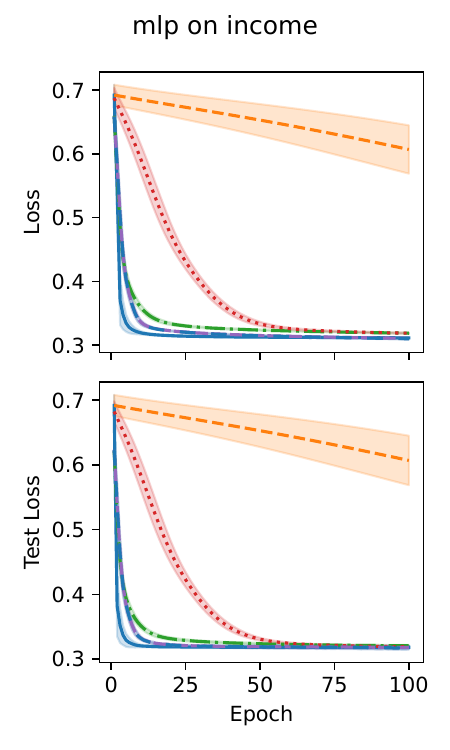}
    \caption{The effect of calibrated noisy gradient covariance as a preconditioner for logistic regression and MLPs on classification tasks for telescope, bank marketing, and adult income datasets.}
    \label{fig:pc_linear_mlp_append}
\end{figure}

\subsection{Ablation with exact GGN blocks}
\label{sec:oracle_ggn}

To assess how curvature-estimation accuracy affects downstream optimisation, we compare our CNGC-based blockwise GD, PC method with an idealised oracle variant. The oracle replaces the estimated curvature blocks with exact GGN blocks computed from the full Jacobian, keeping the block structure and remaining optimisation procedure unchanged. The experimental setup is identical to Figure~\ref{fig:pc_cnn_transf}. This ablation isolates the effect of curvature estimation from the other components of the optimiser. Table~\ref{tab:oracle_test_loss} reports the best test loss achieved, and Table~\ref{tab:oracle_epochs} reports the number of epochs required to reach within 10\% of the best test loss. All results are averaged over 10 random seeds and reported with one standard deviation.

The two variants achieve broadly comparable best test losses, consistent with the similar optimisation trajectories observed in most settings. On CNN--Occupancy, both require an average of 1.10 epochs to reach the threshold. Exact curvature can nevertheless improve convergence, particularly on CNN--Air, where the oracle requires 3.70 epochs compared with 11.00 for CNGC. These results indicate that CNGC retains much of the curvature information useful for this optimisation procedure, although estimation error can affect convergence speed. CNGC obtains this information from aggregate clean and noisy batch gradients without explicitly constructing the Jacobian; the curvature-estimation runtime experiments demonstrate the computational advantage of this approach (see Figures~\ref{fig:est},~\ref{fig:est_appendix}).

\begin{table}[t]
\centering
\caption{Best test loss achieved using CNGC and exact GGN blocks. The results are mean $\pm$ one standard deviation over 10 random seeds.}
\label{tab:oracle_test_loss}
\small
\setlength{\tabcolsep}{4pt}
\resizebox{\linewidth}{!}{%
\begin{tabular}{lccccc}
\hline \noalign{\vskip 2pt}
Optimiser &
\shortstack{MLP\\Telescope} &
\shortstack{CNN\\Occupancy} &
\shortstack{CNN\\Air} &
\shortstack{Transformer\\Occupancy} &
\shortstack{Transformer\\Air} \\
\hline
GD, PC (ours)
& $0.3005 \pm 0.0032$ & $0.1080 \pm 0.0065$
& $0.4053 \pm 0.0200$ & $0.1136 \pm 0.0573$
& $0.1456 \pm 0.0096$ \\
Exact GGN blocks (oracle)
& $0.2981 \pm 0.0040$ & $0.1082 \pm 0.0130$
& $0.3887 \pm 0.0214$ & $0.0918 \pm 0.0058$
& $0.1384 \pm 0.0118$ \\
\hline
\end{tabular}%
}
\end{table}

\begin{table}[t]
\centering
\caption{Number of training epochs required to reach within 10\% of the best test loss for each optimiser and model--dataset combination. $\dagger$ marks an entry for which the threshold was not reached within the training budget for some of the seeds. The results are mean $\pm$ one standard deviation over 10 random seeds.}
\label{tab:oracle_epochs}
\small
\setlength{\tabcolsep}{4pt}
\begin{tabular}{lccccc}
\hline \noalign{\vskip 2pt}
Optimiser &
\shortstack{MLP\\Telescope} &
\shortstack{CNN\\Occupancy} &
\shortstack{CNN\\Air} &
\shortstack{Transformer\\Occupancy} &
\shortstack{Transformer\\Air} \\
\hline
GD, PC (ours)
& $4.50 \pm 0.50$ & $1.10 \pm 0.30$ & $11.00 \pm 3.82$
& $3.70 \pm 5.46^{\dagger}$ & $7.50 \pm 2.29$ \\
Exact GGN blocks (oracle)
& $3.40 \pm 0.80$ & $1.10 \pm 0.30$ & $3.70 \pm 1.10$
& $2.40 \pm 1.62$ & $7.20 \pm 3.16$ \\
\hline
\end{tabular}%
\end{table}

\subsection{CNGC estimator for an early stopping criterion}

\begin{figure}[]
    \centering 
    \includegraphics[width=0.204\linewidth,trim={11 12 10 30},clip]{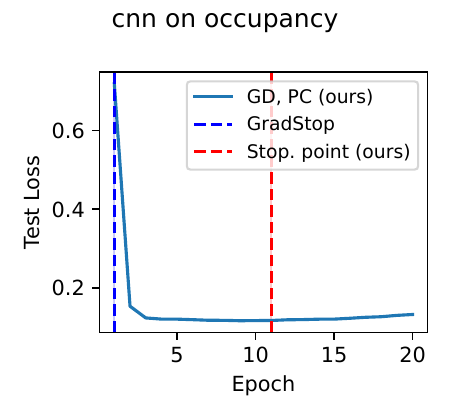}
    \caption{The calculated stopping points on the occupancy detection dataset with GradStop and the proposed criterion with CNGC.}
    \label{fig:early_stop_ind}
\end{figure}

Here, we demonstrate a single run of the early stopping criterion discussed in the main text. The gradient covariance calculated from the clean gradients may result in premature stopping of the method, as shown in Figure~\ref{fig:early_stop_ind}.

To extend this illustrative run, we compare CNGC-based stopping, the blockwise GradStop baseline, and standard validation-loss stopping across the five model--dataset settings used in the optimisation experiments. Validation stopping uses a 90/10 training--validation split, while both gradient-based criteria use the full training set. All three methods stop after three consecutive increases in their respective criterion (patience~3). The CNGC criterion reuses the GGN blocks accumulated during optimisation, requiring no additional curvature estimation.
Table~\ref{tab:early_stopping_comparison} summarises the results. CNGC achieves lower mean test loss than GradStop on four of the five tasks and is broadly competitive with validation-based stopping without requiring a validation split. These results illustrate a further application of the recovered curvature beyond preconditioning.

\begin{table}[h]
\centering
\caption{Test loss obtained with each early-stopping criterion using patience~3. The results are mean $\pm$ one standard deviation over 10 random seeds.}
\label{tab:early_stopping_comparison}
\small
\setlength{\tabcolsep}{4pt}
\resizebox{\linewidth}{!}{%
\begin{tabular}{lccccc}
\hline \noalign{\vskip 2pt}
Stopping criterion &
\shortstack{MLP\\Telescope} &
\shortstack{CNN\\Occupancy} &
\shortstack{CNN\\Air} &
\shortstack{Transformer\\Occupancy} &
\shortstack{Transformer\\Air} \\
\hline
CNGC (ours)
& $0.3075 \pm 0.0034$ & $0.1279 \pm 0.0127$
& $0.4647 \pm 0.0234$ & $0.1223 \pm 0.0580$
& $0.1843 \pm 0.0271$ \\
GradStop
& $0.3056 \pm 0.0039$ & $0.1350 \pm 0.0196$
& $0.4799 \pm 0.0419$ & $0.1227 \pm 0.0610$
& $0.2411 \pm 0.1246$ \\
Validation loss
& $0.3061 \pm 0.0039$ & $0.1273 \pm 0.0197$
& $0.5223 \pm 0.2019$ & $0.1078 \pm 0.0137$
& $0.1646 \pm 0.0096$ \\
\hline
\end{tabular}%
}
\end{table}

\newpage
\section*{NeurIPS Paper Checklist}

\begin{enumerate}

\item {\bf Claims}
    \item[] Question: Do the main claims made in the abstract and introduction accurately reflect the paper's contributions and scope?
    \item[] Answer: \answerYes{} 
    \item[] Justification: The main claims in the abstract and introduction accurately reflect the paper’s scope and contributions. In particular, they match the central claim that variance-calibrated target perturbations recover the GGN from aggregate mini-batch gradients.

\item {\bf Limitations}
    \item[] Question: Does the paper discuss the limitations of the work performed by the authors?
    \item[] Answer: \answerYes{} 
    \item[] Justification: The paper includes a limitations section, which notes that the method recovers the Generalised Gauss--Newton matrix rather than the full Hessian, except in special cases (e.g., GLMs or blockwise ReLU settings). It may incur additional cost due to the extra backward pass required by the centred estimator. It also assumes access to per-example outputs and targets within each minibatch, and in practice often relies on diagonal or block-diagonal approximations to make matrix storage and inversion tractable.

\item {\bf Theory assumptions and proofs}
    \item[] Question: For each theoretical result, does the paper provide the full set of assumptions and a complete (and correct) proof?
    \item[] Answer: \answerYes{} 
    \item[] Justification: The paper states the general assumptions in the Definitions section, and any additional assumptions specific to a given result are stated immediately before that theorem or proposition. Complete proofs for all theoretical results are provided in the appendix.

    \item {\bf Experimental result reproducibility}
    \item[] Question: Does the paper fully disclose all the information needed to reproduce the main experimental results of the paper to the extent that it affects the main claims and/or conclusions of the paper (regardless of whether the code and data are provided or not)?
    \item[] Answer: \answerYes{} 
    \item[] Justification: The paper relies on publicly available datasets for the evaluations easily accessible through UCI repository or OpenML. The experimental artifacts (code, python environment, random seeds, SLURM-scripts, etc.) are provided in a GitHub repository accompanying the paper with the exact commands in README.md file. The main details needed for interpreting the results of are additionally provided in the supplementary material.

\item {\bf Open access to data and code}
    \item[] Question: Does the paper provide open access to the data and code, with sufficient instructions to faithfully reproduce the main experimental results, as described in supplemental material?
    \item[] Answer: \answerYes{} 
    \item[] Justification: The paper relies on publicly available datasets for the evaluations easily accessible through UCI repository or OpenML. The experimental artifacts (code, python environment, random seeds, SLURM-scripts, etc.) are provided in the GitHub repository \url{https://github.com/edahelsinki/cngc}.

\item {\bf Experimental setting/details}
    \item[] Question: Does the paper specify all the training and test details (e.g., data splits, hyperparameters, how they were chosen, type of optimizer) necessary to understand the results?
    \item[] Answer: \answerYes{} 
    \item[] Justification: The supplementary material covers the main experimental details required to understand the results. The full set of details is available via the code provided in the GitHub repository \url{https://github.com/edahelsinki/cngc}.

\item {\bf Experiment statistical significance}
    \item[] Question: Does the paper report error bars suitably and correctly defined or other appropriate information about the statistical significance of the experiments?
    \item[] Answer: \answerYes{} 
    \item[] Justification: All results are reported as the mean with 1-sigma bars over 10 repetitions by varying the random seed.

\item {\bf Experiments compute resources}
    \item[] Question: For each experiment, does the paper provide sufficient information on the computer resources (type of compute workers, memory, time of execution) needed to reproduce the experiments?
    \item[] Answer: \answerYes{} 
    \item[] Justification: The GitHub repository \url{https://github.com/edahelsinki/cngc}  contains a folder with the templates of SLURM scripts used to start the jobs for each experiment. The scripts specify the type of computational resources, memory limits, and execution time limits. All jobs can be run on CPU or GPU and were performed on the internal SLURM-based cluster. The full research project used more computational resources than the experiments reported in the paper (due to preliminary investigation and testing of the hypotheses).
    
\item {\bf Code of ethics}
    \item[] Question: Does the research conducted in the paper conform, in every respect, with the NeurIPS Code of Ethics \url{https://neurips.cc/public/EthicsGuidelines}?
    \item[] Answer: \answerYes{} 
    \item[] Justification: Our research is in full compliance with NeurIPS Code of Ethics. The paper presents theoretical computer science research on publicly available datasets. The paper does not involve research with human subjects.

\item {\bf Broader impacts}
    \item[] Question: Does the paper discuss both potential positive societal impacts and negative societal impacts of the work performed?
    \item[] Answer: \answerYes{} 
    \item[] Justification: The paper is foundational research and is not tied to any specific applications. In practice, the results can be used, e.g., to build a more efficient optimiser for deep learning applications.
    
\item {\bf Safeguards}
    \item[] Question: Does the paper describe safeguards that have been put in place for responsible release of data or models that have a high risk for misuse (e.g., pre-trained language models, image generators, or scraped datasets)?
    \item[] Answer: \answerNA{} 
    \item[] Justification: The paper poses no such risks, the datasets used in the evaluations are publicly available, the paper does not develop or train generative models.

\item {\bf Licenses for existing assets}
    \item[] Question: Are the creators or original owners of assets (e.g., code, data, models), used in the paper, properly credited and are the license and terms of use explicitly mentioned and properly respected?
    \item[] Answer: \answerYes{} 
    \item[] Justification: Our research code and experiments are build with open-source software and active publicly available datasets (accessed through UCI repository or OpenML repository). The original sources of the datasets are cited in the supplementary material.

\item {\bf New assets}
    \item[] Question: Are new assets introduced in the paper well documented and is the documentation provided alongside the assets?
    \item[] Answer: \answerYes{} 
    \item[] Justification: the source code and the additional details are provided in the GitHub repository \url{https://github.com/edahelsinki/cngc}.

\item {\bf Crowdsourcing and research with human subjects}
    \item[] Question: For crowdsourcing experiments and research with human subjects, does the paper include the full text of instructions given to participants and screenshots, if applicable, as well as details about compensation (if any)? 
    \item[] Answer: \answerNA{}  
    \item[] Justification: The paper does not contain research results with human subjects.

\item {\bf Institutional review board (IRB) approvals or equivalent for research with human subjects}
    \item[] Question: Does the paper describe potential risks incurred by study participants, whether such risks were disclosed to the subjects, and whether Institutional Review Board (IRB) approvals (or an equivalent approval/review based on the requirements of your country or institution) were obtained?
    \item[] Answer: \answerNA{} 
    \item[] Justification:  The paper does not contain research results with human subjects.

\item {\bf Declaration of LLM usage}
    \item[] Question: Does the paper describe the usage of LLMs if it is an important, original, or non-standard component of the core methods in this research? Note that if the LLM is used only for writing, editing, or formatting purposes and does \emph{not} impact the core methodology, scientific rigor, or originality of the research, declaration is not required.
    \item[] Answer: \answerYes{} 
    \item[] Justification: We have used LLMs to rapidly produce initial verification of our ideas, in addition to aid in writing, editing, and formatting. The final text, derivations, and experiments, however, are by us.

\end{enumerate}

\end{document}